\documentclass[journal]{IEEEtran}
\usepackage{amsfonts}
\usepackage{eurosym}
\usepackage{cite}
\usepackage{amsmath,bm}
\usepackage{amsmath,graphicx}
\usepackage{amssymb}
\usepackage{tabularx}
\usepackage{multirow}
\usepackage{textcomp}
\usepackage[english]{babel}
\usepackage{listings}
\usepackage{enumitem}
\usepackage{color}
\usepackage{MnSymbol}
\usepackage{algpseudocode}
\usepackage{algorithm}
\usepackage{mathtools}

\ifCLASSINFOpdf
\else
\fi
\makeatletter
\def\endthebibliography{
\def\@noitemerr{\@latex@warning{Empty `thebibliography' environment}}
\endlist
}
\makeatother
\newcommand{\bmu}{\boldsymbol{\mu}}

\def\x{{\mathbf x}}
\def\z{{\mathbf z}}

\def\X{{\mathbf X}}
\def\Z{{\mathbf Z}}
\def\V{{\mathbf V}}
\def\W{{\mathbf W}}

\def\eps{{\varepsilon}}

\def\beq{\begin{equation}}
\def\eeq{\end{equation}}
\def\bal{\begin{aligned}}
\def\eal{\end{aligned}}
\def\beq{\begin{equation}}
\def\eeq{\end{equation}}

\begin{document}

\title{A State-Space Approach to\\ Nonstationary Discriminant Analysis}
%\title{State–Space Discriminant Analysis under\\ Temporal Distribution Shift}

\author{Shuilian Xie, Mahdi Imani, Edward R. Dougherty, and Ulisses M. Braga-Neto 
\thanks{
S Xie, E Dougherty and U Braga-Neto are with the Department of
Electrical and Computer Engineering, Texas A\&M University, and M. Imani is with the Department of Electrical and Computer Engineering at Northeastern University. Email: ulisses@tamu.edu.}}
\maketitle

\begin{abstract}
Classical discriminant analysis assumes identically distributed training data, yet in many applications observations are collected over time and the class-conditional distributions drift. This \emph{population drift} renders stationary classifiers unreliable. We propose a principled, model-based framework that embeds discriminant analysis within state-space models to obtain \emph{nonstationary linear discriminant analysis} (NSLDA) and \emph{nonstationary quadratic discriminant analysis} (NSQDA). For linear–Gaussian dynamics, we adapt Kalman smoothing to handle multiple samples per time step and develop two practical extensions: (i) an expectation–maximization (EM) approach that jointly estimates unknown system parameters, and (ii) a Gaussian mixture model (GMM)–Kalman method that simultaneously recovers unobserved time labels and parameters, a scenario common in practice. To address nonlinear or non-Gaussian drift, we employ particle smoothing to estimate time-varying class centroids, yielding fully nonstationary discriminant rules. Extensive simulations demonstrate consistent improvements over stationary linear discriminant analysis (LDA), quadratic discriminant analysis (QDA), and support vector machine (SVM) baselines, with robustness to noise, missing data, and class imbalance. This paper establishes a unified and data-efficient foundation for discriminant analysis under temporal distribution shift.
\end{abstract}


\IEEEpeerreviewmaketitle

\section{Introduction}
\label{sec:intro}
Many modern learning systems pool observations collected over time. As physical, social, or environmental processes evolve, the underlying distributions drift, violating the i.i.d.\ assumption and degrading stationary classifiers \cite{DevrGyorLugo:96,DudaHartStor:01,Bish:95,lu2019conceptdrift_review,sugiyama2012machine,gama2014conceptdrift,souza2019onlineLDA}. We focus on the regime where the \emph{class-conditional} distributions evolve with time, so the Bayes decision boundary is intrinsically time indexed. Pooling all data averages incompatible regimes, whereas training only on recent data discards useful history and inflates variance; the issue is compounded by irregular sampling, multiple measurements sharing a time index, and noisy or missing timestamps. These realities motivate an explicit temporal model that yields decision rules tailored to a target time while borrowing statistical strength from all observations.

Our goal is to model the temporal evolution of class-conditional structure and produce decision rules targeted to a chosen time. Reactive strategies—change detection, sliding windows, forgetting factors, and ensemble resets—treat time only implicitly~\cite{gama2014conceptdrift,souza2019onlineLDA}; they depend on thresholds and frequent labels, discard informative history, and handle irregular sampling, multiple measurements at the same time index, or noisy/missing timestamps poorly \cite{ditzler2015learning}. A principled alternative is to parameterize the dynamics that govern class moments and priors and to infer them by borrowing statistical strength across the entire horizon, so that the resulting discriminants align with test-time distributions rather than with a pooled average. This setting differs from \emph{covariate shift} \cite{sugiyama2012machine,gama2014conceptdrift,Koh2021WILDS,Yao2022WildTime}, where only the input marginal $p(x)$ changes while $p(y\!\mid\!x)$ remains fixed; here, $p(x\!\mid\!y,t)$ (and possibly class priors) evolves with time, making time-indexed decision boundaries essential.

We adopt a model-based perspective. We posit that the temporal evolution of each class-conditional distribution follows a (possibly nonlinear, non-Gaussian) state-space process~\cite{doucet2001smc,doucet2011tutorial,delmoral2004fk,durbin2012state,SarkkaSvensson2023BFS}. Given observations arriving at assorted times, we infer time-specific class moments by smoothing—borrowing statistical strength across the full horizon—and plug these estimates into Gaussian discriminants to obtain \emph{nonstationary linear discriminant analysis} (NSLDA) and \emph{nonstationary quadratic discriminant analysis} (NSQDA). This yields a classifier at a target time while systematically leveraging all available data.

The linear–Gaussian case is treated first. We derive Kalman-smoother–based estimators that explicitly handle multiple observations at the same time step via sequential updates after a single prediction. To address incomplete metadata, we develop two estimation mechanisms: an expectation–maximization (EM) procedure that jointly estimates unknown system parameters (initial states, dynamics, and process noise) together with latent trajectories, and a Gaussian-mixture–assisted Kalman approach (GMM–Kalman) that simultaneously recovers unknown time labels and parameters. To capture more general temporal dynamics, we also formulate a nonlinear, non-Gaussian variant in which latent class centroids evolve through a flexible state-space model; these trajectories are estimated using particle smoothing (sequential Monte Carlo), and the resulting time-indexed moments define the NSLDA/NSQDA rules.

This framework adapts the decision rule to the current time while retaining all observations for estimation, handles missing data naturally through prediction within the smoother, and remains effective under class imbalance by controlling each class’s contribution during smoothing. Across controlled simulations, the proposed methods improve upon stationary {linear discriminant analysis} (LDA), {quadratic discriminant analysis} (QDA), and {support vector machine} (SVM) baselines and remain robust under high noise, missing observations, and unbalanced training sets.

The main contributions of the paper are as follows: 
\begin{itemize}[leftmargin=*,itemsep=1pt,topsep=2pt]
  \item A unified state-space framework for nonstationary discriminant analysis that produces time-indexed NSLDA/NSQDA via smoothing.
  \item A multi-measurement Kalman smoothing formulation (one prediction, sequential updates) for multiple observations per time step.
  \item Practical estimators under incomplete metadata: EM for unknown system parameters and GMM–Kalman for jointly inferring unknown time labels and parameters.
  \item A nonlinear, non-Gaussian extension using particle smoothing to estimate latent class trajectories and construct the corresponding nonstationary discriminants.
\end{itemize}

The remainder of the paper is organized as follows. Section III introduces the overall framework and notation for nonstationary discriminant analysis. Section IV develops the linear Gaussian model, Kalman smoothing with multiple measurements per time, and practical estimators—EM for unknown parameters and a GMM–Kalman scheme for missing time labels. Section V extends the framework to nonlinear and non-Gaussian dynamics using particle smoothing and derives the corresponding time-indexed NSLDA/NSQDA rules. Section VI presents simulations covering noise, missing data, and class imbalance, with comparisons against LDA/QDA/SVM baselines. Finally, Section VII includes the conclusion.

\section{Nonstationary Discriminant Analysis}
In a nonstationary classification problem, there is a feature vector $\X_k \in R^d$ with class label $Y_k \in \{0,1,\ldots,c-1\}$ at each discrete time $k=0,1,\ldots$ (e.g., obtained by sampling an underlying continuous-time process). Let $f^j_{k}(\x)=p(\x\mid Y_k=j)$ denote the class-conditional density (probability mass function in the discrete case) and let $\pi^j_k=\mathbb{P}(Y_k=j)$ be the prior probability of class $j$ at time $k$, with $\sum_{j=0}^{c-1}\pi_k^j=1$. Population drift is the joint evolution of $\{f^j_{k},\pi^j_k\}$ with $k$.

Under $0$--$1$ loss, the (time-$k$) risk of a classifier $\psi$ is $R_k(\psi)=\mathbb{P}\{\psi(\X_k)\neq Y_k\}$, and the Bayes rule at time $k$ is
\begin{equation}
\psi_k^*(\x)\,=\,\operatornamewithlimits{argmax}_{j=0,1,\ldots ,c-1}D_k^{*,j}\!(\x)\,,
\label{eq:opt}
\end{equation}
where the corresponding discriminants are
\begin{equation}
D_k^{*,j}\!(\x) \,=\, \log \pi^j_k + \log f^j_k(\x)\,,
\label{eq-optdisc}
\end{equation}
for $j=0,1,\ldots,c-1$ and $k=0,1,\ldots$. The Bayes error may be written as $\eps_k^* = \sum_{j=0}^{c-1} \pi_k^j\,\mathbb{P}\!\big(\psi_k^*(\X_k)\neq j \mid Y_k=j\big)$ and is minimal among all rules at time $k$. In the stationary special case ($f_k^j \equiv f^j$ and $\pi_k^j \equiv \pi^j$), a single optimal classifier $\psi^*$ and error $\eps^*$ do not depend on time.

In practice, $\{f_k^j,\pi_k^j\}$ are unknown, and one designs sample-based rules from time-labeled data $S_n = \bigcup_{k=0}^T\bigcup_{j=0}^{c-1} S_k^j$, $S_k^j = \{\X_{k,i}^j:\, i=1,\ldots,n_k^j\}$, $n=\sum_{k=0}^{T}\sum_{j=0}^{c-1} n_k^j$. Typically, $T$ is the current time and the goal is a classifier $\psi_{n,T}$; in retrospective studies, a $\psi_{n,k}$ for any $k\le T$ may be desired. Forecasting to $k>T$ is outside our scope. Time labels may be fully observed, partially observed, or missing; in the latter case they must be estimated.

A stationary baseline pools all samples and learns a single rule, ignoring time. A naive nonstationary baseline trains a separate rule using only $\{S_k^j\}_j$ at each time $k$, discarding data at other times and suffering when $n_k^j$ is small. A model-based alternative is to describe how $\{f_k^j,\pi_k^j\}$ evolve with $k$ and to design a classifier for each time $k$ using \emph{all} the data via smoothing—thereby borrowing statistical strength across time while adapting the decision rule to the target time.

\paragraph*{Gaussian specialization}
Assume the class-conditional densities are multivariate Gaussian~\cite{duda2001pattern}:
\[
   f^j_{k}(\x\mid Y_k=j) \,=\, \frac{1}{\sqrt{(2\pi )^{d}\det(\Sigma_k^j)}}
   \exp\!\left(-\frac{1}{2}\,\big\|\x-\bmu_k^j\big\|^2_{\Sigma^j_k}\right),
\]
where $\|\mathbf{v}\|^2_M \,=\, \mathbf{v}^T M^{-1}\mathbf{v}$. Ignoring additive constants common to all classes, the Bayes discriminants in~\eqref{eq-optdisc} become
\begin{equation}
  D_k^{*,j}\!(\x)\,=\, \log \pi^{j}_k -\frac{1}{2}\log \big|\Sigma^{j}_k\big| \;-\; \frac{1}{2}\,\big\|\x-\bmu_k^j\big\|^2_{\Sigma^j_k}\,,
\label{eq:discri3}
\end{equation}
and, under homoskedasticity at time $k$ (i.e., $\Sigma^j_k \equiv \Sigma_k$ across $j$),
\begin{equation}
  D_k^{*,j}\!(\x)\,=\, \log \pi^{j}_k \;-\; \frac{1}{2}\,\big\|\x-\bmu_k^j\big\|^2_{\Sigma_k}\,.
\label{eq:discri3L}
\end{equation}

\paragraph*{Plug-in QDA/LDA at time $k$}
Given estimators $\hat{\pi}_k^j$, $\hat{\bmu}^j_k$, and $\hat{\Sigma}^j_k$ from the data $S_n$, the plug-in \emph{quadratic} discriminants are
\begin{equation}
  D_{n,k}^{{\rm QDA},j}\!(\x)\,=\, \log \hat{\pi}^{j}_k \;-\; \frac{1}{2}\log \big|\hat{\Sigma}^{j}_k\big| \;-\; \frac{1}{2}\,\big\|\x-\hat{\bmu}_k^j\big\|^2_{\hat{\Sigma}^j_k}\,,
\label{eq:QDAdisc}
\end{equation}
and, under homoskedasticity at time $k$, the \emph{linear} discriminants are obtained by replacing $\hat{\Sigma}^j_k$ with the within-time pooled covariance
\begin{equation}
  \hat{\Sigma}_k \,=\, \frac{\sum_{j=0}^{c-1}(n_k^j-1)\,\hat{\Sigma}_k^j}{\sum_{j=0}^{c-1}(n_k^j-1)}\,,
\label{eq:pooled}
\end{equation}
giving
\begin{equation}
  D_{n,k}^{{\rm LDA},j}\!(\x)\,=\, \log \hat{\pi}^{j}_k \;-\; \frac{1}{2}\,\big\|\x-\hat{\bmu}_k^j\big\|^2_{\hat{\Sigma}_k}\,.
\label{eq:LDAdisc}
\end{equation}
The corresponding sample-based classifiers are
\begin{equation}
\psi^{\rm QDA}_{n,k}(\x)\,=\,\operatornamewithlimits{argmax}%
_{j=0,1,\ldots ,c-1}D_{n,k}^{{\rm QDA},j}\!(\x)\,,
\label{eq:QDAcl}
\end{equation}
\begin{equation}
\psi^{\rm LDA}_{n,k}(\x)\,=\,\operatornamewithlimits{argmax}%
_{j=0,1,\ldots ,c-1}D_{n,k}^{{\rm LDA},j}\!(\x)\,,
\label{eq:LDAcl}
\end{equation}
yielding piecewise quadratic and linear decision boundaries, respectively. In the general (heteroskedastic) case the time-indexed rule is quadratic—nonstationary QDA (NSQDA); under homoskedasticity it reduces to a linear rule—nonstationary LDA (NSLDA).

The quality of these plug-in rules depends on estimator accuracy. Priors $\hat{\pi}_k^j$ may come from domain knowledge or empirical class proportions when time-labeled data are sufficient; when neither is reliable, a no-information default $\hat{\pi}_k^j=1/c$ is common. For small $n_k^j$, covariance estimators may be regularized (e.g., diagonal or shrinkage variants) to improve stability. A naive time-slice approach estimates $(\hat{\bmu}_k^j,\hat{\Sigma}_k^j)$ using only $S_k^j$, which is statistically inefficient unless $n_k^j$ is large. In this paper, we instead estimate \emph{time-indexed} moments by smoothing across \emph{all} times using state-space models—Kalman smoothing in the linear–Gaussian setting (with support for multiple observations per time step and EM/GMM procedures for unknown parameters and unknown time labels; see Section~\ref{sec:linear}) and particle smoothing in the nonlinear/non-Gaussian setting (Section~\ref{sec:nonlinear}). The resulting NSQDA/NSLDA rules adapt the decision boundary to the target time $k$ while borrowing information across the full horizon.

\section{Linear Drift Model}
\label{sec:linear}

\subsection{State-Space Model}
\label{sub:linear-drifts} 

We assume the following linear–Gaussian state-space generative model for the feature vector $\X_{k}^j\!\in\!R^d$ of class $j$ at time $k$~\cite{sarkka2013bayesian,durbin2012state,SarkkaSvensson2023BFS}:
\begin{equation}
\begin{aligned} 
  \Z_{k+1}^j & \,=\, A^j \Z_{k}^j \,+\, \W_k^j,\\[0.5ex]
  \X_{k}^j   & \,=\, \Z_k^j \,+\, \V_{k}^j \,,
\end{aligned}  
\label{eq:state-linear}
\end{equation}
for $k=0,1,\ldots$, where $\Z_0^j \sim \mathcal{N}(\bmu_0^j,K^j)$. The noise processes $\{\W_k^{j}\}$ and $\{\V_{k}^{j}\}$ are zero-mean Gaussian, temporally white, mutually independent, and independent of $\Z_0^j$, with covariance matrices $Q^j$ and $R^j$, respectively; processes are independent across classes. The matrix $A^j$ governs the linear drift of the latent state (e.g., contraction/rotation), while $\W_k^j$ induces stochastic translation. Under mild stability conditions (e.g., bounded horizon or $\rho(A^j)\le 1$), second moments remain finite.

\begin{figure}[t!] 
	\centering 
	\includegraphics[scale=0.45]{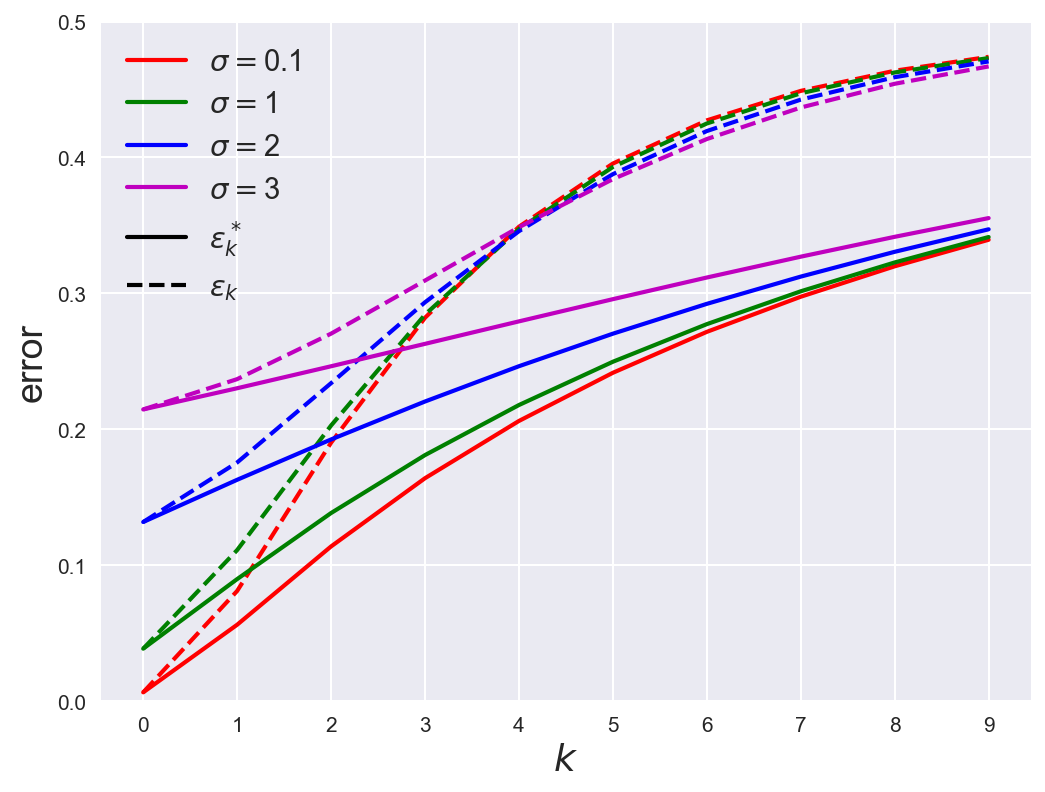} \vspace{-2ex}
	\caption{Classification error vs.\ time in the linear model example with $a = 0.9$, $\mu_0^0=5$, $\mu_0^1=10$, $\sigma_Q^2=\sigma_R^2=1$, and varying initial-state standard deviation $\sigma$. The error of the frozen \emph{time-0} rule increases much faster than that of the optimal time-$k$ rule.} 
	\label{fig:example} 
\end{figure}

\vspace{1ex}
\noindent
\emph{Example.} Consider the two-class univariate model
\begin{equation}
\begin{aligned} 
  Z_{k+1}^j & \,=\, a\, Z_{k}^j \,+\, W_k^j \,,\\[0.5ex]
  X_k^j     & \,=\, Z_k^j \,+\, V_k^j \,,
\end{aligned}  
\label{eq:example0}
\end{equation}
for $j\in\{0,1\}$ and $k=0,1,\ldots$, where $0<a<1$, $Z_0^j \sim \mathcal{N}(\mu_0^j,\sigma^2)$, and $W_{k}^{j}$ and $V_{k}^{j}$ have constant variances $\sigma_Q^2$ and $\sigma_R^2$, respectively. This models noisy measurements of quantities that gradually decay over time (e.g., due to diffusive or corrosive processes). Then $X_k^j \sim \mathcal{N}(\mu_k^j,\sigma_k^2)$ with
\begin{equation}
\begin{aligned} 
  \mu_k^j &\,=\, \mathbb{E}[Z_k^j] \,=\, a^k\mu_0^j, \\[0.25ex]
  \sigma_k^2 &\,=\, \mathrm{Var}(Z_k^j) + \sigma_R^2 \,=\, a^{2k} \sigma^2 + \frac{1-a^{2k}}{1-a^2}\,\sigma_Q^2 + \sigma_R^2.
\end{aligned}  
\label{eq:example1}
\end{equation}
Assuming equal class priors and $\mu_0^1 > \mu_0^0$, it follows from (\ref{eq:opt}) and (\ref{eq:discri3}) that the optimal classifier at time $k$ is the threshold rule
\[
  \psi_k^*(x) \,=\, 
  \begin{cases}
  1, & x > \dfrac{\mu_k^0+\mu_k^1}{2} \;=\; a^k \dfrac{\mu_0^0+\mu_0^1}{2},\\[1ex]
  0, & \text{otherwise.}
  \end{cases}
\]
The corresponding Bayes error at time $k$ is~\cite{Brag:20}
\[
  \eps^*_k \,=\, \Phi\!\left(-\frac{a^k}{2}\,\frac{\mu_0^1-\mu_0^0}{\sigma_k}\right),
\]
where $\Phi$ is the standard normal cdf. As $k$ increases, the class means drift toward each other and $\eps_k^*$ increases. If, instead, one naively applies the \emph{time-0} optimal classifier at all times, the error at time $k$ becomes
\[
  \eps_k \,=\, \tfrac{1}{2}\,\Phi\!\left(-\frac{(2a^k\!-\!1)\mu_0^1 - \mu_0^0}{2\,\sigma_k}\right)
           \;+\; \tfrac{1}{2}\,\Phi\!\left(-\frac{\mu_0^1-(2a^k\!-\!1)\mu_0^0}{2\,\sigma_k}\right),
\]
which grows faster than $\eps_k^*$, as illustrated in Figure~\ref{fig:example}.

\subsection{Parameter Estimation with Modified Kalman Smoother}
\label{sub:kal} 
From (\ref{eq:state-linear}), we have that $\X_k^j \sim {\cal N}(\bmu_k^j,\Sigma_k^j)$, where
\begin{equation}
\begin{aligned} 
  \bmu_k^j &\,=\, E[\Z_k^j] \,=\, (A^j)^k\bmu_0^j\,, \\
   \Sigma_k^j & \,=\, {\rm Cov}(\Z_k^j) + R^j \\[-1ex]
   &\!\!\!\!=\, (A^j)^k K^j\big((A^j)^T\big)^k \;+\; \sum_{i=0}^{k-1} (A^j)^i Q^j \big((A^j)^T\big)^i \;+\; R^j\,.
\end{aligned}  
\label{eq:general}
\end{equation}
Notice that if $A^j \equiv A$, $K^j \equiv K$, $Q^j \equiv Q$, then ${\rm Cov}(\Z_k^j)$ does not depend on $j$. If in addition $R^j \equiv R$ (a reasonable assumption), then the model is homoskedastic at each separate time $k$, i.e., $\Sigma^j_k \equiv \Sigma_k$, which leads to an LDA discriminant (see the previous section).

Based on (\ref{eq:general}), and given data $\{S_0^j,S_1^j,\ldots,S_T^j\}$ for class $j$, we propose time–$k$ plug-in estimators
\begin{equation}
\begin{aligned} 
  \hat{\bmu}_k^j &\,=\, E[\Z_k^j \mid S_0^j,S_1^j,\ldots,S_T^j] \;=\; \hat{\z}^j_{k|T}\,,\\[0.75ex]
  \hat{\Sigma}_k^j &\,=\, P^j_{k|k-1} \;+\; \hat{R}^j\,,
\end{aligned}  
\label{eq:genest}
\end{equation}
where $P^j_{k|k-1}$ is the \emph{prediction} covariance (the prior variance of $\Z_k^j$ at time $k$) from the forward pass, and $\hat{R}^j$ is an estimate of the measurement noise covariance. Using $P^j_{k|k-1}$ (rather than the smoothed posterior $P^j_{k|T}$) matches the population covariance ${\rm Cov}(\Z_k^j)$ that appears in $\Sigma_k^j={\rm Cov}(\Z_k^j)+R^j$.

To estimate $R^j$, let $N^j=\sum_{k=0}^{T} n_k^j$ be the number of samples for class $j$. A residual-based, bias-corrected estimator is
\begin{equation}
  \hat{R}^j \,=\, \frac{1}{N^j}\!
  \sum_{k=0}^{T}\sum_{i=1}^{n_k^j} \big(\X_{k,i}^j-\hat{\z}^j_{k|T}\big)\big(\X_{k,i}^j-\hat{\z}^j_{k|T}\big)^{T}
  \;-\; \frac{1}{N^j}\!\sum_{k=0}^{T} n_k^j\, P^j_{k|T}.
\label{eq:Rest}
\end{equation}
Since $E[\Z_k^j \mid S_0^j,\ldots,S_T^j]$ is the MMSE estimator, the class-mean estimator is unbiased:
\[
  E[\hat{\bmu}_k^j] \,=\, E\!\left[E[\Z_k^j \mid S_0^j,\ldots,S_T^j]\right]
  \,=\, E[\Z_k^j] \,=\, E[\X_k^j] \,=\, \bmu_k^j\,.
\]

Define
\[
  \hat{\z}^j_{k|l} \,=\, E[\Z_k^j \mid S_0^j,\ldots,S_l^j]\,, 
  \qquad
  P^j_{k|l} \,=\, {\rm Cov}(\Z_k^j \mid S_0^j,\ldots,S_l^j)\,,
\]
for $k,l=0,\ldots,T$. The modified multiple-update Kalman smoother computes $\hat{\z}^j_{k|T}$ and $P^j_{k|T}$ based on the entire data. As stated in (\ref{eq:genest}), the population parameters are then estimated by $\hat{\bmu}_k^j=\hat{\z}^j_{k|T}$ and $\hat{\Sigma}_k^j=P^j_{k|k-1}+\hat{R}^j$. The smoother consists of a forward and a backward pass; because multiple measurements may arrive at the same time step, one prediction is followed by \emph{sequential} updates (optimal for multiple measurements~\cite{willner1976kalman}). Assuming for now that $\bmu_0^j$, $K^j$, $A^j$, and $Q^j$ are known and all samples are time-labeled (both assumptions are relaxed later), the recursions are~\cite{haykin2001kalman}:

\vspace{1ex}
\noindent
\underline{Forward pass:}

\begin{itemize}[leftmargin=+.1in]
\item Initialization: 
\begin{equation*}
\hat{\z}_{0|0}^{j}\,=\, \bmu_0^j, \qquad P^j_{0|0} \,=\, K^j.
\end{equation*}

\item For $k=1,\ldots,T$, do: 
\begin{itemize}[leftmargin=+.05in]
\item Prediction:
\begin{align*}
\hat{\z}_{k|k-1}^{j}& \,=\,A^j\,\hat{\z}_{k-1|k-1}^{j}\,, \\
P_{k|k-1}^{j}& \,=\,A^{j}P_{k-1|k-1}^{j}(A^j)^{T}\,+ \,Q^{j}.
\end{align*}

\item Sequential updates (identity measurement):
\begin{align*}
&\hat{\z}_{k|k,0}^{j} \,=\, \hat{\z}_{k|k-1}^{j},\qquad  P^j_{k|k,0} \,=\, P^j_{k|k-1}\,,\\
&\text{for } i=1,\ldots,n_{k}^{j}: \\
&\hspace{1em} S_{k,i}^j \,=\, P_{k|k,i-1}^{j}+ R^{j}, \qquad
K_{k,i}^{j} \,=\,P_{k|k,i-1}^{j}\,(S_{k,i}^j)^{-1}, \\
&\hspace{1em} \hat{\z}_{k|k,i}^{j} \,=\,\hat{\z} _{k|k,i-1}^{j}\,+\,K_{k,i}^{j}\,\big( \X_{k,i}^{j}\,-\,\hat{\z}_{k|k,i-1}^{j}\big) \,, \\
&\hspace{1em} P_{k|k,i}^{j} \,=\,\big(I\!-\!K_{k,i}^{j}\big)\,P_{k|k,i-1}^{j}\,\big(I\!-\!K_{k,i}^{j}\big)^{T}
\,+\, K_{k,i}^{j}\,R^{j}\,(K_{k,i}^{j})^{T}\,,
\end{align*}
and set $\hat{\z}_{k|k}^{j}=\hat{\z}_{k|k,n_{k}^j}^{j}$, $P^j_{k|k}=P^j_{k|k,n_{k}^j}$.
\end{itemize}
\end{itemize}

\noindent
Here, $K_{k,i}^{j}$ is the \emph{Kalman gain} for observation $\X_{k,i}^j$ at time $k$ and $S_{k,i}^j$ is the innovation covariance.

\vspace{1ex}
\noindent
\underline{Backward pass:}

\begin{itemize}
\item For {$k=T-1,\ldots,0$}, do: 
\begin{align*}
L_{k}^{j}& \,=\,P_{k|k}^{j}(A^{j})^{T}\,\big(P_{k+1|k}^{j}\big)^{-1}\,, \\
\hat{\z}_{k|T}^{j}& \,=\,\hat{\z}_{k|k}^{j}\,+\,L_{k}^{j}\big( \hat{\z} _{k+1|T}^{j}\,-\,\hat{\z}_{k+1|k}^{j}\big)\,, \\
P_{k|T}^{j}& \,=\,P_{k|k}^{j}\,+\,L_{k}^{j}\big(
P_{k+1|T}^{j}\,-\,P_{k+1|k}^{j}\big) \,(L_{k}^{j})^{T}\,,
\end{align*}
\end{itemize}
where $L_{k}^{j}$ is the \emph{smoother gain}. Its form reveals a correction of the forward estimate using only forward-pass quantities. In addition, the smoothed mean $\hat{\z}_{k|T}^{j}$ does not depend on the smoothed covariance $P_{k|T}^{j}$; to obtain $\hat{\z}_{k|T}^{j}$, only the forward state estimates and the smoother gain need be stored.

\subsection{Adaptive Filter for Systems with Unknown Parameters}
\label{sub:EM-and-GMM}
When the parameters in equations~(\ref{eq:state-linear}) are only partially unknown, the Kalman smoother—originally developed for known linear–Gaussian state–space models—must be embedded in an \emph{adaptive} procedure that jointly identifies model parameters and estimates latent states. For example, when the initial state mean ${\mbox{\boldmath $\pi$}}_1^j$, initial state covariance $V_1^j$, dynamics matrix $A^{j}$, or process–noise covariance $Q^{j}$ in~(\ref{eq:state-linear}) are unknown, we use Expectation–Maximization (EM)~\cite{ghahramani1996parameter,digalakis1993ml,dempster1977em,shumway1982em,ljung1999system,ljung2010perspectives} in combination with the (modified) Kalman smoother of Section~\ref{sub:kal} to estimate the unknown parameters together with the time–indexed class means. When the measurement time labels are missing, we combine a Gaussian mixture model (GMM)~\cite{bouman1997cluster,duda2001pattern,cappe2005ihmm} with Kalman filtering/smoothing to infer the labels and any remaining unknown parameters~\cite{sarkka2013bayesian,cappe2005ihmm}. Throughout this section we adopt $C^j=I$ unless stated and allow multiple observations at a given time index, i.e., $\{\x_{k,i}^j\}_{i=1}^{n_k^j}$ at time $k$.

\subsubsection{EM-based Kalman Smoother}
\label{subsub:EM-based}
Ordinary maximum–likelihood estimation~\cite{shumway2017time} maximizes the \emph{incomplete} log–likelihood (with $\z_{1:T}^j$ marginalized out). Here both the state sequence and parameters are unknown; EM instead maximizes the expected \emph{complete} log–likelihood, which is quadratic for the linear–Gaussian model. Using~(\ref{eq:state-linear}), the complete–data log–density for class $j$ is, up to an additive constant,
\begin{equation}
\begin{aligned}
\label{eq:fun_Q2}
&\log p\!\left(\z^j_{1:T},\x^j_{1:T}\right)\\
&\propto
-\frac{1}{2}\sum_{k=1}^{T}\sum_{i=1}^{n_k^j}\!\left(\x_{k,i}^j - C^j \z_k^j\right)^{\!T}\!\left(R^j\right)^{-1}\!\left(\x_{k,i}^j - C^j \z_k^j\right)
-\frac{1}{2}\sum_{k=1}^{T}\!n_k^j \log|R^j| \\
&\quad
-\frac{1}{2}\sum_{k=2}^{T}\!\left(\z_k^j - A^j \z_{k-1}^j\right)^{\!T}\!\left(Q^j\right)^{-1}\!\left(\z_k^j - A^j \z_{k-1}^j\right)
-\frac{T-1}{2}\log|Q^j| \\
&\quad
-\frac{1}{2}\left(\z_1^j - {\mbox{\boldmath $\pi$}}_1^j\right)^{\!T}\!\left(V_1^{j}\right)^{-1}\!\left(\z_1^j - {\mbox{\boldmath $\pi$}}_1^j\right)
-\frac{1}{2}\log|V_1^j|\,.
\end{aligned}
\end{equation}

Given current parameters $\hat{\boldsymbol\theta}_n^j$, the E–step forms
\begin{equation}
\label{eq:fun_Q1}
\mathcal{Q}\!\left(\boldsymbol\theta^j,\hat{\boldsymbol\theta}_n^j\right)
= \mathbb{E}\!\left[\log p\!\left(\z^j_{1:T},\x^j_{1:T}\right)\,\middle|\,\x^j_{1:T},\hat{\boldsymbol\theta}_n^j\right],
\end{equation}
where the expectation is with respect to the smoothing distribution produced by the (modified) Kalman smoother tuned to $\hat{\boldsymbol\theta}_n^j$ (one prediction followed by $n_k^j$ sequential updates at each time $k$). For $k=1,\ldots,T$, define
\begin{equation}
\begin{aligned}
\hat{\z}_{k|T}^{\,j} &\equiv \mathbb{E}[\z_k^j \mid \x^j_{1:T}], \\
V_{k|T}^{\,j} &\equiv \mathrm{Cov}(\z_k^j \mid \x^j_{1:T}), \\
V_{k,k-1|T}^{\,j} &\equiv \mathrm{Cov}(\z_k^j,\z_{k-1}^j\mid \x^j_{1:T}),
\end{aligned}
\end{equation}
and the second moments
\begin{equation}
\begin{aligned}
P_{k|T}^{\,j} &\equiv \mathbb{E}[\z_k^j \z_k^{jT}\mid \x^j_{1:T}] = V_{k|T}^{\,j}+\hat{\z}_{k|T}^{\,j}\hat{\z}_{k|T}^{\,jT}, \\
P_{k,k-1|T}^{\,j} &\equiv \mathbb{E}[\z_k^j \z_{k-1}^{jT}\mid \x^j_{1:T}] = V_{k,k-1|T}^{\,j}+\hat{\z}_{k|T}^{\,j}\hat{\z}_{k-1|T}^{\,jT}.
\end{aligned}
\end{equation}
The cross–covariances $V_{k,k-1|T}^{\,j}$ are obtained by the standard Rauch–Tung–Striebel cross–covariance recursion; see, e.g.,~\cite{shumway2017time}.

The M–step maximizes~(\ref{eq:fun_Q1}) in closed form,
\begin{equation}
\label{eq:M-step}
\hat{\boldsymbol\theta}_{n+1}^j=\operatornamewithlimits{argmax}_{\boldsymbol\theta^j}\;
\mathcal{Q}\!\left(\boldsymbol\theta^j,\hat{\boldsymbol\theta}_n^j\right),
\end{equation}
yielding
\begin{itemize}
\item \textbf{State dynamics matrix:}
\begin{equation}
\label{eq:EM_A}
\hat{A}_{n+1}^{\,j}
= \left(\sum_{k=2}^{T} P_{k,k-1|T}^{\,j}\right)
  \left(\sum_{k=2}^{T} P_{k-1|T}^{\,j}\right)^{-1}.
\end{equation}

\item \textbf{Process–noise covariance:}
\begin{equation}
\begin{aligned}
\label{eq:EM_Q}
\hat{Q}_{n+1}^{\,j}
&= \frac{1}{T-1}\sum_{k=2}^{T}
\Big(P_{k|T}^{\,j}
- \hat{A}_{n+1}^{\,j} P_{k,k-1|T}^{\,jT}
\\
&- P_{k,k-1|T}^{\,j} \hat{A}_{n+1}^{\,jT}
+ \hat{A}_{n+1}^{\,j} P_{k-1|T}^{\,j} \hat{A}_{n+1}^{\,jT}\Big).
\end{aligned}
\end{equation}

\item \textbf{Initial state mean/covariance:}
\begin{equation}
\label{eq:EM_piV1}
\hat{{\mbox{\boldmath $\pi$}}}_{1,n+1}^{\,j} = \hat{\z}_{1|T}^{\,j}, \qquad
\hat{V}_{1,n+1}^{\,j} = P_{1|T}^{\,j} - \hat{\z}_{1|T}^{\,j}\hat{\z}_{1|T}^{\,jT}.
\end{equation}

\item \textbf{(Optional) Measurement–noise covariance (with $C^j=I$):}
\begin{equation}
\begin{aligned}
\label{eq:EM_R}
\hat{R}_{n+1}^{\,j}
&= \frac{1}{N^j}\sum_{k=1}^{T}\sum_{i=1}^{n_k^j}
\Big(\x_{k,i}^j-\hat{\z}_{k|T}^{\,j}\Big)\Big(\x_{k,i}^j-\hat{\z}_{k|T}^{\,j}\Big)^{\!T}
+ \frac{1}{N^j}\sum_{k=1}^{T} n_k^j\, V_{k|T}^{\,j},\\
N^j &\triangleq \sum_{k=1}^{T} n_k^j.
\end{aligned}
\end{equation}
\end{itemize}

E– and M–steps are iterated until a prescribed tolerance (e.g., $\max$ absolute parameter change $<\epsilon$) is met. EM monotonically increases the observed–data log–likelihood and converges to a stationary point. In finite samples, if $\hat Q^{\,j}$, $\hat R^{\,j}$, or $\hat V_{1}^{\,j}$ is not positive semidefinite, we project onto the PSD cone. After convergence, the smoother tuned to $\hat{\boldsymbol\theta}^{\,j}$ returns the time–indexed observation–space moments used downstream (with $C^j=I$): $\hat{\bmu}_k^j=\hat{\z}_{k|T}^{\,j}$ and $\hat{\Sigma}_k^j = V_{k|T}^{\,j}+\hat R^{\,j}$.

\vspace{2mm}
\subsubsection{GMM-based Kalman Smoother}
\label{subsub:GMM-based}
We now consider missing time labels. Let $t_i^j\in\{1,\ldots,T\}$ denote the (latent) time index of $\x_i^j\in S_n^j$, and let $\boldsymbol\theta^j$ collect the (possibly unknown) model parameters with prior $p(\boldsymbol\theta^j)$ and parameter space $\Theta^j$. A joint MAP estimator seeks
\begin{equation}
\label{eq:opt2}
\big(\boldsymbol\theta^{*\,j},\{t_i^{*\,j}\}_{i=1}^{n}\big)
=\operatornamewithlimits{argmax}_{\hat{\boldsymbol\theta}\in\Theta^j,\;\{t_i\}\subset\{1,\ldots,T\}}
\;\log p(\hat{\boldsymbol\theta})\;+\;\log p\!\big(S_n^j \mid \hat{\boldsymbol\theta},\{t_i\}\big),
\end{equation}
which is generally hard to optimize jointly. We adopt a pragmatic two–stage scheme based on a $T$–component GMM whose components correspond to time indices. For a candidate $\hat{\boldsymbol\theta}=\{A^j,Q^j,{\mbox{\boldmath $\pi$}}_1^j,V_1^j,R^j\}$ and $C^j=I$, the implied observation–space moments are
\begin{equation}
\begin{aligned}
\label{eq:mus_sigmas}
\z^{\mathrm{GMM},j}_{1,\hat{\boldsymbol\theta}} &= {\mbox{\boldmath $\pi$}}_1^j, \qquad
\Sigma^{\mathrm{GMM},j}_{1,\hat{\boldsymbol\theta}} = V_1^j + R^j,\\
\z^{\mathrm{GMM},j}_{k,\hat{\boldsymbol\theta}} &= A^j\,\z^{\mathrm{GMM},j}_{k-1,\hat{\boldsymbol\theta}},\\
\Sigma^{\mathrm{GMM},j}_{k,\hat{\boldsymbol\theta}} &= A^j\,\Sigma^{\mathrm{GMM},j}_{k-1,\hat{\boldsymbol\theta}}(A^j)^{T} + Q^j + R^j,\qquad k\ge 2,
\end{aligned}
\end{equation}
which parameterize Gaussian components $\{\mathcal{N}(\z^{\mathrm{GMM},j}_{k,\hat{\boldsymbol\theta}},\Sigma^{\mathrm{GMM},j}_{k,\hat{\boldsymbol\theta}})\}_{k=1}^{T}$. Assign each $\x_i^j\in S_n^j$ a hard time label by maximum component likelihood:
\begin{equation}
\label{eq:stp1}
t_i^{\,j,\hat{\boldsymbol\theta}}
= \operatornamewithlimits{argmax}_{k\in\{1,\ldots,T\}}
\ \mathcal{N}\!\big(\x_i^j;\,\z^{\mathrm{GMM},j}_{k,\hat{\boldsymbol\theta}},\,\Sigma^{\mathrm{GMM},j}_{k,\hat{\boldsymbol\theta}}\big).
\end{equation}
Given the assigned labels $\{t_i^{\,j,\hat{\boldsymbol\theta}}\}$, we score parameters by the (unnormalized) posterior
\begin{equation}
\label{eq:pos}
p\!\left(\hat{\boldsymbol\theta}\mid \{t_i^{\,j,\hat{\boldsymbol\theta}}\},S_n^j\right)
\propto p(\hat{\boldsymbol\theta})\times p\!\big(S_n^j\mid \hat{\boldsymbol\theta},\{t_i^{\,j,\hat{\boldsymbol\theta}}\}\big),
\end{equation}
where the likelihood on the right is computed by a Kalman filter that, for each occupied time step $k$, performs \emph{one prediction} followed by $n_k^j$ \emph{sequential updates} (to handle multiple measurements at the same time). Over a finite candidate set $\Theta=\{\hat{\boldsymbol\theta}^{(m)}\}_{m=1}^{M}$ (e.g., coarse EM initializations or a stochastic search such as MCMC~\cite{ghahramani1996parameter,shumway1982approach}), we select
\begin{equation}
\label{eq:maxpos}
\hat{\boldsymbol\theta}^{\,*,j}
=\operatornamewithlimits{argmax}_{\hat{\boldsymbol\theta}\in\Theta}
\Big\{\log p(\hat{\boldsymbol\theta})+\log p\!\big(S_n^j\mid \hat{\boldsymbol\theta},\{t_i^{\,j,\hat{\boldsymbol\theta}}\}\big)\Big\}.
\end{equation}
Finally, with $(\hat{\boldsymbol\theta}^{\,*,j},\{t_i^{\,*,j}\})$ fixed, we run the (modified) fixed–interval smoother to obtain the time–indexed class moments $\{\hat{\bmu}_k^j,\hat{\Sigma}_k^j\}_{k=1}^{T}$, where $\hat{\bmu}_k^j=\hat{\z}_{k|T}^{\,j}$ and $\hat{\Sigma}_k^j=V_{k|T}^{\,j}+\hat R^{\,j}$.

\subsection{EM-based and GMM-based Nonstationary Linear Discriminant Analysis}
\label{sub:EM-GMM-NSLDA} 

After computation of the class-conditional moments at time steps
$k=1,\ldots,T$ by the Kalman smoother (KS) tuned to the inferred parameters
$\hat{\boldsymbol\theta}^{*,j}$, we classify at each time $k$ using the
(time-varying) Gaussian models in the \emph{observation space}. With
$C^j=I$, define
\[
\hat{\bmu}_k^{\,j}\triangleq \hat{\z}_{k|T}^{\,j},\qquad
\hat{\Sigma}_k^{\,j}\triangleq P_{k|T}^{\,j}+ \hat R^{\,j},
\]
where $\hat{\z}_{k|T}^{\,j}$ and $P_{k|T}^{\,j}$ are the KS mean and state covariance, and
$\hat R^{\,j}$ is the (possibly estimated) measurement-noise covariance for class $j$.
(For a general $C^j$, replace $\hat{\Sigma}_k^{\,j}$ by $C^j P_{k|T}^{\,j} C^{jT} + \hat R^{\,j}$.)

Let $\pi_k^j$ denote the class prior at time $k$. Then the (time-indexed)
\emph{quadratic} discriminant (QDA) comparing class $1$ to class $0$ can be written as
\begin{equation}
\begin{aligned}
D_k(\x) &= \x^T E_k \x + F_k^T \x + G_k, 
\end{aligned}
\label{eq:discri1-linear}
\end{equation}
with
\begin{equation}
\begin{aligned}
E_k \;&=\; -\tfrac{1}{2}\!\left((\hat{\Sigma}_k^{\,1})^{-1}-(\hat{\Sigma}_k^{\,0})^{-1}\right),\\
F_k \;&=\; (\hat{\Sigma}_k^{\,1})^{-1}\hat{\bmu}_k^{\,1}
          \;-\;(\hat{\Sigma}_k^{\,0})^{-1}\hat{\bmu}_k^{\,0},\\
G_k \;&=\; -\tfrac{1}{2}\,\hat{\bmu}_k^{\,1T}(\hat{\Sigma}_k^{\,1})^{-1}\hat{\bmu}_k^{\,1}
           \;+\;\tfrac{1}{2}\,\hat{\bmu}_k^{\,0T}(\hat{\Sigma}_k^{\,0})^{-1}\hat{\bmu}_k^{\,0}
           \;-\;\tfrac{1}{2}\log\frac{|\hat{\Sigma}_k^{\,1}|}{|\hat{\Sigma}_k^{\,0}|}\,,
\end{aligned}
\label{eq:discri2-linear}
\end{equation}
and the Bayes decision with priors is
\begin{equation}
\psi_k(\x)=
\begin{cases}
1, & D_k(\x) + \log\!\dfrac{\pi_k^1}{\pi_k^0} \;\geq\; 0,\\[1ex]
0, & \text{otherwise.}
\end{cases}
\label{eq:discri3-linear}
\end{equation}
In the \emph{homoskedastic} special case $\hat{\Sigma}_k^{\,0}=\hat{\Sigma}_k^{\,1}\equiv \hat{\Sigma}_k$,
the rule reduces to the (time-varying) \emph{linear} discriminant (NSLDA):
\begin{equation}
\begin{aligned}
w_k \;&=\; \hat{\Sigma}_k^{-1}\!\left(\hat{\bmu}_k^{\,1}-\hat{\bmu}_k^{\,0}\right),\\
b_k \;&=\; -\tfrac{1}{2}\!\left(\hat{\bmu}_k^{\,1T}\hat{\Sigma}_k^{-1}\hat{\bmu}_k^{\,1}
          -\hat{\bmu}_k^{\,0T}\hat{\Sigma}_k^{-1}\hat{\bmu}_k^{\,0}\right)
          + \log\!\frac{\pi_k^1}{\pi_k^0},
\end{aligned}
\end{equation}
and decide class $1$ iff $w_k^T\x + b_k \ge 0$.
\emph{Numerical note:} to avoid instability when inverting ill-conditioned $\hat{\Sigma}_k^{\,j}$,
we use shrinkage $\hat{\Sigma}_k^{\,j}\leftarrow \hat{\Sigma}_k^{\,j}+\lambda I$ if needed.

\vspace{1ex}
The whole process of the proposed EM-based and GMM-based nonstationary discriminants
is summarized in Algorithm~\ref{alg:EM-NSLDA} and Algorithm~\ref{alg:GMM-NSLDA}, respectively.

\begin{algorithm}
\caption{EM-based Linear/Quadratic Nonstationary Discriminant Analysis}
\begin{algorithmic}[1]
\small
\State \textbf{For} $j \in \{0,1,\ldots\}$ \textbf{do}:
\begin{enumerate}
\item[--] \textbf{EM process:}
  \begin{enumerate}
  \item[--] Initialize $\hat{\boldsymbol\theta}_0^{\,j}$; set $n\!=\!0$.
  \item[] \textbf{Repeat}
    \begin{enumerate}
    \item[--] \emph{E-step:} Run the Kalman smoother tuned to $\hat{\boldsymbol\theta}_n^{\,j}$ to obtain
               $\{\hat{\z}_{k|T}^{\,j}, P_{k|T}^{\,j}\}_{k=1}^{T}$.
    \item[--] \emph{M-step:} Update $\boldsymbol\theta^{\,j}$ using the closed-form EM updates
               \eqref{eq:EM_A}--\eqref{eq:EM_R}.
    \item[--] $n \leftarrow n+1$.
    \end{enumerate}
  \item[] \textbf{Until} $\max\!\big\|\hat{\boldsymbol\theta}_n^{\,j}-\hat{\boldsymbol\theta}_{n-1}^{\,j}\big\|_\infty < \epsilon$.
  \end{enumerate}
\item[--] Set $\hat{\boldsymbol\theta}^{\,j} \leftarrow \hat{\boldsymbol\theta}_n^{\,j}$.
\item[--] Run the KS tuned to $\hat{\boldsymbol\theta}^{\,j}$ and form
          $\hat{\bmu}_k^{\,j}\!=\!\hat{\z}_{k|T}^{\,j}$ and $\hat{\Sigma}_k^{\,j}\!=\!P_{k|T}^{\,j}+\hat R^{\,j}$.
\end{enumerate}
\State \textbf{EndFor}
\State \textbf{Classification at each time $k=1,\ldots,T$:}
Use \eqref{eq:discri1-linear}--\eqref{eq:discri3-linear};
if $\hat{\Sigma}_k^{\,0}=\hat{\Sigma}_k^{\,1}$, the linear NSLDA form applies.
\end{algorithmic}\label{alg:EM-NSLDA}
\end{algorithm}

\begin{algorithm}[h]\small
\caption{GMM-based Nonstationary Linear/Quadratic Discriminant Analysis}
\begin{algorithmic}[1]
\For{$\hat{\boldsymbol\theta} \in \{\hat{\boldsymbol\theta}^{(1)},\ldots,\hat{\boldsymbol\theta}^{(M)}\}$}
  \For{$j\in\{0,1\}$}
    \State $\z^{\mathrm{GMM},j}_{0,\hat{\boldsymbol\theta}}=\z_{0,\hat{\boldsymbol\theta}}^{\,j}$,\quad
           $\Sigma^{\mathrm{GMM},j}_{0,\hat{\boldsymbol\theta}}=P_{0,\hat{\boldsymbol\theta}}^{\,j}$
    \For{$k=1,2,\ldots,T$}
      \State $\z^{\mathrm{GMM},j}_{k,\hat{\boldsymbol\theta}}=A_{\hat{\boldsymbol\theta}}^{\,j}\,
             \z^{\mathrm{GMM},j}_{k-1,\hat{\boldsymbol\theta}}$
      \State $\Sigma^{\mathrm{GMM},j}_{k,\hat{\boldsymbol\theta}}
              =A_{\hat{\boldsymbol\theta}}^{\,j}\,\Sigma^{\mathrm{GMM},j}_{k-1,\hat{\boldsymbol\theta}}\!
               \left(A_{\hat{\boldsymbol\theta}}^{\,j}\right)^{\!T}
               +Q_{\hat{\boldsymbol\theta}}^{\,j}+R_{\hat{\boldsymbol\theta}}^{\,j}$
    \EndFor
    \State \emph{Hard assignment of time labels:}
           $t_i^{\,j,\hat{\boldsymbol\theta}}=\arg\max_{k\in\{1,\ldots,T\}}
           \mathcal{N}\!\big(\X_i^{\,j};\,\z^{\mathrm{GMM},j}_{k,\hat{\boldsymbol\theta}},
           \Sigma^{\mathrm{GMM},j}_{k,\hat{\boldsymbol\theta}}\big)$ for $i=1,\ldots,n$
  \EndFor
  \State $T^{\hat{\boldsymbol\theta}}=\max\big(\{t_i^{\,0,\hat{\boldsymbol\theta}}\}_{i=1}^{n}
         \cup \{t_i^{\,1,\hat{\boldsymbol\theta}}\}_{i=1}^{n}\big)$
  \State \emph{KF scoring:} for $j\in\{0,1\}$, run a Kalman filter tuned to $\hat{\boldsymbol\theta}$ with
         one prediction and $n_k^j$ sequential updates at each occupied $k$ to obtain
         $\{\hat{\z}_{k|k,\hat{\boldsymbol\theta}}^{\,j},P_{k|k,\hat{\boldsymbol\theta}}^{\,j}\}_{k=1}^{T^{\hat{\boldsymbol\theta}}}$
         and the log-likelihood $L^j(\hat{\boldsymbol\theta})$.
\EndFor
\State \emph{Time labels and parameter selection (per class $j$):}
\[
\hat{\boldsymbol\theta}^{*,j}=\arg\max_{\hat{\boldsymbol\theta}\in\{\hat{\boldsymbol\theta}^{(m)}\}}
\left\{\log p(\hat{\boldsymbol\theta}) + L^j(\hat{\boldsymbol\theta})\right\}.
\]
\State Set $T=T^{\hat{\boldsymbol\theta}^*}$ and run the KS tuned to $\hat{\boldsymbol\theta}^{*,j}$ to get
       $\{\hat{\z}_{k|T}^{\,j},P_{k|T}^{\,j}\}_{k=1}^{T}$, and form
       $\hat{\bmu}_k^{\,j}=\hat{\z}_{k|T}^{\,j}$, $\hat{\Sigma}_k^{\,j}=P_{k|T}^{\,j}+\hat R^{\,j}$.
\State \textbf{Classification at each time $k=1,\ldots,T$:}
use \eqref{eq:discri1-linear}--\eqref{eq:discri3-linear}.
(If an equal-covariance assumption is adopted, use the linear NSLDA form with
$\hat{\Sigma}_k\!\triangleq\!\tfrac{1}{2}(\hat{\Sigma}_k^{\,0}+\hat{\Sigma}_k^{\,1})$.)
\end{algorithmic}\label{alg:GMM-NSLDA}
\end{algorithm}

\section{Non-Linear State Space Model Approach}
\label{sec:nonlinear}
\subsection{Nonlinear Drift Model}

\label{sub:nonlinear-model} 
We relax linearity and Gaussianity by modeling the temporal evolution of the class-conditional distributions with a general nonlinear state-space model~\cite{doucet2001smc,doucet2011tutorial,delmoral2004fk,Naesseth2019Elements}. Training a separate classifier at each time point is often impractical—data may be scarce, missing, or noisy—so we estimate the evolving class moments by smoothing over time with sequential Monte Carlo (SMC) methods~\cite{doucet2001sequential,kantas2015particle}. We adopt a particle smoother~\cite{hurzeler1998monte,douc2011smoothing_hmm,Naesseth2019Elements} that we modify to accommodate multiple observations per time index, under the standard assumption that observations are conditionally independent given the latent state at that time. Once time-indexed moments are estimated, we plug them into LDA/QDA discriminants to obtain nonstationary classifiers~\cite{duda2001pattern}. We now formalize the nonlinear drift model.
%In this paper, the restrictive assumptions such as linearity of the state-space model or Gaussianity of the noise process, are omitted by modeling the evolution of the class conditional distributions using general state-space models. Training classifiers at any given time point using the available data from various classes might not be practical or lead to poor classification performance. This is due to factors such as data limitation, missing data, or large noise in the data. To overcome these difficulties, we propose using sequential Monte-Carlo (SMC) techniques~\cite{doucet2001sequential,kantas2015particle} for efficient estimation of the class-conditional distributions via a finite set of particles. This is achieved by the use of a particle smoother technique~\cite{hurzeler1998monte}, modified here to handle multiple data at each time point. Upon representing the underlying process of the class-conditional distributions, any discriminant analysis classifiers can be employed for decision making using the sets of particles at different time points. We need to define the nonlinear drifts in class-conditional distributions.

In a multi-class nonstationary problem with $c$ classes and $T$ time points, we assume that the centroid of each class is a latent variable that evolves in time according to the following nonlinear model: 
\begin{equation}
\mathbf{z}_{k}^{j}\,=\,\mathbf{f}_{k}^{j}\left( \mathbf{z}_{k-1}^{j},\mathbf{%
	w}_{k}^{j}\right) ,  \label{eq:state}
\end{equation}
for $j=0,1,\ldots ,c-1$ and $k=1,\ldots ,T$, where $\mathbf{f}_{k}^{j}$ is an arbitrary nonlinear function governing the evolution of class $j$ and $ \mathbf{w}_{k}^{j}$ defines an i.i.d.\ transition noise process, which is independent of the $\mathbf{z}_{k}^{j}$ process. The initial states $\mathbf{
	z}_{0}^{j}$ are generated from given starting initial distributions. We assume ${\mathbf{w}k^j}$ are i.i.d. with known density and independent of ${\mathbf{v}{k,i}^j}$ and of the initial state $\mathbf{z}_0^j$. The initial states $\mathbf{z}_0^j$ are drawn from $p(\mathbf{z}_0^j)$.

For notational simplicity, we partition the training data into $c \times T$ subsamples 
\begin{equation}
S_{k}^j \,=\, \{\mathbf{x}_{k,1}^j,\ldots,\mathbf{x}_{k,n_k^j}^j\},
\end{equation}
for $j=0,\ldots,c-1$ and $k=1,\ldots,T$, where $n_k^j$ are the sample sizes for each class $j$ at time $k$, adding up to the total sample size $n$. Nothing is assumed in this paper about the sampling mechanism; e.g., for fixed $k$, $n_k^j$ could be a random variable or a fixed experimental design parameter (future work will examine the sampling issue). The data are assumed to satisfy the following general observation model: 
\begin{equation}
\mathbf{x}_{k,i}^j \,=\, \mathbf{h}_k^j(\mathbf{z}_{k}^j,\mathbf{v}%
_{k,i}^j)\,,  \label{eq:obs}
\end{equation}
for $i=1,\ldots,n_k^j$, $j=0,1,\ldots,c-1$ and $k=1,\ldots,T$, where $\mathbf{h}_k^j$ is an arbitrary nonlinear function mapping the latent variables to the observable data and $\mathbf{v}_{k,i}^j$ defines an i.i.d.\ observation noise process, which is independent of the $\mathbf{z}_{k}^j$ process.

In this paper, the ultimate goal is to develop a framework for nonstationary classification when the class-conditional distribution is represented by (\ref{eq:state}) and (\ref{eq:obs}). Due to the general nonlinearity and non-Gaussianity, we use SMC-based smoothing (Section~\ref{sub:particle-smoother}) to estimate ${\mathbf{z}_k^j}$ and time-indexed moments. For clarity of exposition in Section~\ref{sub:SMC-NSLDA}, we specialize $h_k^j$ to an additive Gaussian form; the SMC machinery itself remains applicable to general $h_k^j$.

%Due to the nonlinearity of the state process and non-Gaussianity of the state process noise, in Section \ref{sub:particle-smoother}, we propose using sequential Monte-Carlo (SMC) for estimating the class-conditional distributions. 

\subsection{Particle Smoother for Nonstationary Classification Model Inference}

\label{sub:particle-smoother} 
Our goal here is to estimate the latent trajectories $\{\mathbf z_k^j\}$ given all data $S_{1:T}^j$ for each class $j$. We first present smoothing when the parameter vector $\boldsymbol\theta^j$ (governing $\mathbf f_k^j,\mathbf h_k^j$ and the noise laws) is fixed; when some components of $\boldsymbol\theta^j$ are unknown we update them via a particle–EM scheme in Section~\ref{sub:para-est-ps}. The resulting smoothed moments are used to build the time–indexed classifier in Section~\ref{sub:SMC-NSLDA}.
%Our ultimate goal is to developed a framework for nonstationary classification based on the model described in the previous section \ref{sub:nonlinear-model}. For that purpose, the primary task is to estimate the latent variables $\mathbf{z}_k^j$, but it may also be necessary to estimate the noise parameters. In this paper, we will assume that all noise parameters are known and focus on the estimation of the latent variables given the data. In the next section, we build a classification rule using the results of this section.

Several inference methods exist in the literature that can handle the
nonlinearity and non-Gaussianity of the NSC model~\cite%
{kitagawa1996monte,kantas2015particle}. In this paper, we propose a method
that is based on the particle smoother in~\cite{hurzeler1998monte}, with a
suitable modification to handle multiple independent data at each time
point. For each class, the smoother consists of forward and backward
processes. In the forward process a particle filter algorithm is run to
compute the forward particles and weights characterizing the filtering
distributions. The approximate smoothed distribution is computed by running
a backward process for correcting the filtering weights~\cite{godsill2004monte,briers2010smoothing,fearnhead2010linear}. We describe each of
these steps next.

\subsubsection{Forward Process}
We use the auxiliary particle filter (APF)~\cite{pitt1999}. Let
$S_{1:k}^j = S_1^j \cup \cdots \cup S_k^j$ be all data for class $j$ up to time $k$.
Given particles $\{ \tilde{\mathbf z}_{k-1,r}^j, w_{k-1,r}^j \}_{r=1}^N$ approximating
$p(\mathbf z_{k-1}^j \mid S_{1:k-1}^j)$, the APF computes
first–stage (look–ahead) weights
\[
v_{k,r}^j \;\propto\; p(S_k^j \mid \nu_{k,r}^j)\, w_{k-1,r}^j,
\]
where $\nu_{k,r}^j$ is a deterministic characteristic of the one–step predictive,
e.g., the conditional mean of $\mathbf z_k^j$ given $\tilde{\mathbf z}_{k-1,r}^j$.
We resample ancestor indices $\{\zeta_{k,i}\}_{i=1}^N \sim \mathrm{Cat}(\tilde v_{k,1}^j,\ldots,\tilde v_{k,N}^j)$ and propagate
\[
\tilde{\mathbf z}_{k,i}^j \sim q_k(\,\cdot \mid \tilde{\mathbf z}_{k-1,\zeta_{k,i}}^j\,),
\]
with proposal $q_k$. We use the prior as proposal, $q_k(\mathbf z_k^j \mid \tilde{\mathbf z}_{k-1,\zeta}^j)=p(\mathbf z_k^j \mid \tilde{\mathbf z}_{k-1,\zeta}^j)$,
yielding second–stage weights
\[
w_{k,i}^j \;\propto\; \frac{p(S_k^j \mid \tilde{\mathbf z}_{k,i}^j)}{p(S_k^j \mid \nu_{k,\zeta_{k,i}}^j)}
\;=\; \prod_{r=1}^{n_k^j}\frac{p(\mathbf x_{k,r}^j \mid \tilde{\mathbf z}_{k,i}^j)}
                                     {p(\mathbf x_{k,r}^j \mid \nu_{k,\zeta_{k,i}}^j)}\,.
\]
We compute all products in log–space to avoid underflow and normalize the weights.
Iterating for $k=1{:}T$ yields $\{ \tilde{\mathbf z}_{0:T,i}^j, w_{0:T,i}^j \}_{i=1}^N$.

\subsubsection{Backward Process}
We use fixed–interval smoothing on the particle system (FFBSm).
Initialize $w_{T\mid T,i}^j \propto w_{T,i}^j$.
For $k=T-1,\ldots,0$, update
\[
w_{k\mid T,i}^j \;\propto\; w_{k,i}^j
\sum_{s=1}^N 
\frac{p(\tilde{\mathbf z}_{k+1,s}^j \mid \tilde{\mathbf z}_{k,i}^j)\, w_{k+1\mid T,s}^j}
     {\sum_{\ell=1}^N p(\tilde{\mathbf z}_{k+1,s}^j \mid \tilde{\mathbf z}_{k,\ell}^j)\, w_{k,\ell}^j}\,,
\]
followed by normalization over $i=1{:}N$. This has $O(N^2)$ cost per time $k$; 
for very large $N$ one may instead use backward–simulation smoothing to draw 
smoothed trajectories in $O(N)$ expected time.

% The backward process is based on the following equation: 
% \begin{equation}
% \begin{aligned} p&(\mathbf{z}^j_{k}\mid S^j_{1:T})\\
% &=\int_{\mathbf{z}^j_{k+1}}p(\mathbf{z}^j_{k}\mid
% \mathbf{z}^j_{k+1},S^j_{1:T})\,p(\mathbf{z}^j_{k+1}\mid S^j_{1:T})\,d{\bf
% 	z}^j_{k+1} \notag \\ &=\int_{\mathbf{z}^j_{k+1}}p(\mathbf{z}^j_{k}\mid
% \mathbf{z}^j_{k+1},S^j_{1:k})\,p(\mathbf{z}^j_{k+1}\mid S^j_{1:T})\,d{\bf
% 	z}^j_{k+1} \notag \\
% &=\int_{\mathbf{z}^j_{k+1}}\!\!\!\!\frac{p(\mathbf{z}^j_{k+1}\mid
% 	\mathbf{z}^j _{k})p(\mathbf{z}^j_{k}\mid S^j_{1:k})p(\mathbf{z}^j _{k+1}\mid
% 	S^j_{1:T})}{p(\mathbf{z}^j_{k+1}\mid S^j _{1:k})}\,d{\bf z}^j_{k+1}\,,
% \end{aligned}  \label{eq:Smoo1}
% \end{equation}%
% where $k<T$ and $p(\mathbf{z}_{k+1}^{j}\mid S_{1:T}^{j})$ is the smoothed
% distribution at time step $k+1$. The smoothed weights $w_{T|T,i}^{j}$ are
% just the forward weights $w_{T,i}^{j}$ at the end of the time interval. The
% smoothed weights at time $k<T$ can be obtained recursively: 
% \begin{equation}
% w_{k|T,i}^{j}\,=\,w_{k,i}^{j}\,\sum_{i=1}^{N}\,\frac{p(\tilde{\mathbf{z}}%
% 	_{k+1,i}^{j}\mid \tilde{\mathbf{z}}_{k,i}^{j})\,w_{k+1,i}^{j}}{%
% 	\sum_{l=1}^{N}\,p(\tilde{\mathbf{z}}_{k+1,i}^{j}\mid \tilde{\mathbf{z}}%
% 	_{k,i}^{j})\,w_{k,l}^{j}}\,,  \label{eq:Smoo2}
% \end{equation}%
% for $k=T-1,\ldots ,1$, $i=1,\ldots ,N$.

\subsection{Parameter Estimation via Particle Smoothing}
\label{sub:para-est-ps}
The difficulty of estimating nonlinear systems is widely discussed in \cite{ljung2003bode, ljung2010perspectives,imani2017maximum,schon2011system,ljung2010perspectives}. In this section, we consider some parameters $\mathbf{\theta}$ in Eq. (\ref{eq:state}) and Eq. (\ref{eq:obs}) are unknown, then the nonlinear state space model becomes:
\begin{equation}
\begin{aligned}
\mathbf{z}_{k}^{j}\,&=\,\mathbf{f}_{k}^{j}\left( \mathbf{z}_{k-1}^{j},\mathbf{%
	w}_{k}^{j}, \mathbf{\theta}^j\right) \\
\mathbf{x}_{k,i}^j \,&=\, \mathbf{h}_k^j(\mathbf{z}_{k}^j,\mathbf{v}%
_{k,i}^j, \mathbf{\theta}^j)\,,
\end{aligned}
\end{equation}
In \cite{schon2011system}, a Maximum Likelihood framework is employed and an Expectation Maximization (EM) algorithm is derived to solve the parameter estimation problem in a general class of nonlinear dynamic systems. The EM algorithm obtains a sequence of parameter estimates $\{\theta_{\left(n\right)}^j; n=0,1,...\}$ for class $j$. Given the current estimate $\theta_{n}^j$, the algorithm obtain the next estimate $\theta_{n+1}^j$ in the sequence by computing (E-step) the function (see \cite{schon2011system} for details):

\begin{equation}
\begin{aligned}
{\mathcal{Q}} (\mathbf{\theta}^j,\mathbf{\theta}_{\left(n\right)}^j) &= \int \text{log}\, p_{\theta^{j}} \left( \mathbf{z}^j_{0:T}, S^j_{1:T} \right) p_{\theta_{\left(n\right)}^{j}} \left( \mathbf{z}^j_{0:T} | S^j_{1:T} \right) d\mathbf{z}^j_{0:T} \\
&= I_1 \left(\mathbf{\theta}^j,\mathbf{\theta}_{\left(n\right)}^j \right) + I_2 \left(\mathbf{\theta}^j,\mathbf{\theta}_{\left(n\right)}^j \right) + I_3 \left(\mathbf{\theta}^j,\mathbf{\theta}_{\left(n\right)}^j \right)
\end{aligned}
\end{equation}
where
\begin{align}\label{eq: eq_I}
I_1(\boldsymbol\theta^j,\boldsymbol\theta_{(n)}^j)
&= \int \log p_{\boldsymbol\theta^j}(\mathbf z_0^j)\;
            p_{\boldsymbol\theta_{(n)}^j}(\mathbf z_0^j \mid S_{1:T}^j)\, d\mathbf z_0^j, \\
I_2(\boldsymbol\theta^j,\boldsymbol\theta_{(n)}^j)
&= \sum_{k=0}^{T-1} \iint \log p_{\boldsymbol\theta^j}(\mathbf z_{k+1}^j \mid \mathbf z_k^j)\;\\
           &\qquad\qquad p_{\boldsymbol\theta_{(n)}^j}(\mathbf z_{k+1}^j,\mathbf z_k^j \mid S_{1:T}^j)\, d\mathbf z_k^j d\mathbf z_{k+1}^j, \\
I_3(\boldsymbol\theta^j,\boldsymbol\theta_{(n)}^j)
&= \sum_{k=1}^T \int \log p_{\boldsymbol\theta^j}(S_k^j \mid \mathbf z_k^j)\;
          p_{\boldsymbol\theta_{(n)}^j}(\mathbf z_k^j \mid S_{1:T}^j)\, d\mathbf z_k^j.
\end{align}

% \begin{equation} \label{eq: eq_I1}
% \begin{aligned}
% I_1 \left(\mathbf{\theta}^j,\mathbf{\theta}_{\left(n\right)}^j \right)  = \int \text{log}\, p_{\theta^{j}} \left( \mathbf{z}^j_{0}\right) p_{\theta_{\left(n\right)}^{j}} \left( \mathbf{z}^j_{0} | S^j_{1:T} \right) d\mathbf{z}^j_{0}
% \end{aligned}
% \end{equation}

% \begin{equation} \label{eq: eq_I2}
% \begin{aligned}
% I_2 \left(\mathbf{\theta}^j,\mathbf{\theta}_{\left(n\right)}^j \right)  =  & \sum_{k = 1}^{T}\iint \text{log}\, p_{\theta^{j}} \left( \mathbf{z}^j_{k+1} | z^j_{k}\right) \\
% &p_{\theta_{\left(n\right)}^{j}} \left( \mathbf{z}^j_{k+1}, z^j_{k} | S^j_{1:T} \right) d\mathbf{z}^j_{k}d\mathbf{z}^j_{k+1}
% \end{aligned}
% \end{equation}

% \begin{equation} \label{eq: eq_I3}
% \begin{aligned}
% I_3 \left(\mathbf{\theta}^j,\mathbf{\theta}_{\left(n\right)}^j \right)  =  \sum_{k=1}^{T}\int \text{log}\, p_{\theta^{j}} \left( \mathbf{x}^j_{k} | \mathbf{z}^j_{k}\right) p_{\theta_{\left(n\right)}^{j}} \left( \mathbf{z}^j_{k} | S^j_{1:T} \right) d\mathbf{z}^j_{k}
% \end{aligned}
% \end{equation}
% and then maximizing (M-step) this function:
% \begin{equation}  \label{eq:M-step-nonlinear}
% \mathbf{\theta}_{n+1}^j = \operatornamewithlimits{argmax}_{\mathbf{%
% 		\theta}^j} {\mathcal{Q}}(\mathbf{\theta}^j,\mathbf{\theta}_{\left(n\right)}^j)\,.
% \end{equation}

We develop our EM algorithm based on the particle smoothing shown in section \ref{sub:particle-smoother}, which contains forward and backward steps. Given the particles and weights $\{\tilde{\mathbf{z}}_{k,i}^{j},w_{k|T,i}^{j}\}_{i=1}^{N}$  obtained by running the forward backward filter tuned to parameter $\theta^j_{\left(n\right)}$, equations (\ref{eq: eq_I}) can be approximated as \cite{olsson2008smc,andrieu2010pmcmc,kantas2015particle}:

\begin{equation} \label{eq: eq_I1_appx}
\begin{aligned}
I_1 \left(\mathbf{\theta}^j,\mathbf{\theta}_{\left(n\right)}^j \right)  = \sum_{i=1}^{N} w_{1|T,i}^{j} \text{log} \, p_{\theta^{j}} \left(\tilde{\mathbf{z}}_{1,i}^{j} \right)
\end{aligned}
\end{equation}

\begin{equation} \label{eq: eq_I2_appx}
\begin{aligned}
I_2 \left(\mathbf{\theta}^j,\mathbf{\theta}_{\left(n\right)}^j \right)  =  \sum_{k=1}^{T-1}  \sum_{i=1}^{N} \sum_{s=1}^{N} w_{k|T,is}^{j} \text{log} \, p_{\theta^{j}} \left( \tilde{\mathbf{z}}_{k+1,i}^{j} | \tilde{\mathbf{z}}_{k,i}^{j} \right)
\end{aligned}
\end{equation}

\begin{equation} \label{eq: eq_I3_appx}
\begin{aligned}
I_3 \left(\mathbf{\theta}^j,\mathbf{\theta}_{\left(n\right)}^j \right)  = \sum_{k=1}^{T}  \sum_{i=1}^{N} w_{k|T,i}^{j} \text{log} \, p_{\theta^{j}} 
\left( S_k^j | \tilde{\mathbf{z}}_{k,i}^{j} \right)
\end{aligned}
\end{equation}

Given $\boldsymbol\theta_{(n)}^j$, we compute $I_1,I_2,I_3$ from the smoothed particle system
and maximize $\mathcal Q(\boldsymbol\theta^j,\boldsymbol\theta_{(n)}^j)$ to obtain $\boldsymbol\theta_{(n+1)}^j$.
We use gradient ascent with backtracking line search and project any covariance parameters onto the PSD cone.
We iterate until the relative change in $\mathcal Q$ or in $\boldsymbol\theta^j$ falls below a threshold.
Finally, we rerun the particle smoother at $\boldsymbol\theta^{\mathrm{ML}}$ to report smoothed state estimates.

%The steps of this EM adaptive filter are as follows \cite{imani2018particle}. Initially, N particles and their weights $\{\tilde{\mathbf{z}}_{k,i}^{j},w_{k|T,i}^{j}\}_{i=1}^{N}$ are obtained using the forward backward step developed in auxiliary particle filter (APF) tuned to a initial parameter guess $\theta_{0}^j$ to compute ${\mathcal{Q}} (\mathbf{\theta}^j,\hat{\mathbf{\theta}}_n^j)$ (E-step). The aforementioned gradient-descent procedure is applied to find the best parameter $\theta_{1}^j$ that maximizes  with $\theta_{0}^j$ fixed (M-Step). The obtained parameter vector is set as the parameter for the particle smoother for the next run, and the process continues until there is no significant change in parameter estimates between two consecutive steps, yielding the final parameter estimate $\theta^{ML}$. Then the smoothed state estimates can be obtained by performing a particle smoothing tuned to parameter $\theta^{ML}$.

\subsection{SMC-Based Nonstationary Discriminant Analysis}

\label{sub:SMC-NSLDA} The particles and weights calculated with the particle
smoother in the previous section, together with the information about the
observational model in (\ref{eq:obs}), can be used to approximate the
class-conditional densities $p(\x_{k}\mid y=j)$ for each class $j$
at each time $k$, $j=0,1,\ldots ,c-1$ and $k=1,\ldots ,T$.

In this paper, we will assume a specific case of model (\ref{eq:obs}): 
\begin{equation}
\mathbf{x}_{k,i}^{j}\,=\,\mathbf{z}_{k}^{j}\,+\,\mathbf{v}_{k,i}^{j}\,,
\label{eq:obs2}
\end{equation}%
where $\mathbf{v}_{k,i}^{j}\sim {\mathcal{N}}(\mathbf{0},R^{j})$ is i.i.d.\
zero-mean Gaussian noise with known covariance matrix $R^{j}$, for $%
i=0,1,\ldots ,n_{k}^{j}$, $j=0,1,\ldots ,c-1$, and $k=1,\ldots ,T$.

From (\ref{eq:obs2}), the first and second moments of $p(\mathbf{x}_k \mid y
= j)$ are given by 
\begin{equation}
\begin{aligned} \z_k^j & \,=\, E[{\bf x}_k \mid y = j]
\,=\, E[{\bf z}_k^j] \\ \Sigma_k^j & \,=\, \Sigma_{\z_k}^j + R^j
\end{aligned}
\end{equation}
for $j=0,1,\ldots,c-1$ and $k=1,\ldots,T$.

Using the particles and weights $\{\tilde{\mathbf{z}}%
_{k,i}^{j},w_{k|T,i}^{j}\}_{i=1}^{N}$ calculated previously leads to the
following approximations: 
\begin{equation}
\begin{aligned} \hat{\z}_k^j & \,=\,
\sum_{i=1}^N\,\tilde{\z}^j_{k,i}\,w^j_{k|T,i}\,,\\ \hat{\Sigma}_k^j & \,=\,
\frac{N}{N-1}\sum_{i=1}^N\,w^j_{k|T,i}\left(\tilde{\mathbf{z}}^j_{k,i}-
\hat{\z}_k^{j}\right)\left(\tilde{\mathbf{z}}^j_{k,i}-
\hat{\z}_k^{j}\right)^T \!+\, R^j \end{aligned}  \label{eq:SMCest}
\end{equation}%
for $j=0,1,\ldots ,c-1$ and $k=1,\ldots ,T$.

\emph{Quadratic Discriminant Analysis} (QDA)~\cite{DudaHartStor:01} relies
on general estimators $\hat{\z}^{j}$ and $\hat{\Sigma}%
^{j}$, for $j=0,\ldots ,c-1$, of the first and second moments of the
class-conditional densities to obtain a classifier. Given the \emph{%
discriminants} (from the theory of optimal classification with Gaussian
class-conditional densities) 
\begin{equation}
D_{\mathrm{QDA}}^{j}\!(\mathbf{x})=\log \pi ^{j}-\frac{1}{2}\log |\hat{\Sigma%
}^{j}|-\frac{1}{2}(\mathbf{x}-\hat{\z}^{j})^{T}(\hat{%
\Sigma}^{j})^{-1}(\mathbf{x}-\hat{\z}^{j}),
\label{eq:discri33}
\end{equation}%
where $\pi ^{j}=P(y=j)$ is the class prior probability or estimate of the
same, the general QDA classifier is given by 
\begin{equation}
\psi _{\mathrm{QDA}}(\mathbf{x})\,=\,\operatornamewithlimits{argmax}%
_{j=0,1,\ldots ,c-1}D_{\mathrm{QDA}}^{j}\!(\mathbf{x})\,.
\end{equation}

On the other hand, \emph{Linear Discriminant Analysis} (LDA)~\cite%
{DudaHartStor:01} is based on the discriminants: 
\begin{equation}
D_{\mathrm{LDA}}^j\!(\mathbf{x}) = \log \pi^j -\frac{1}{2}(\mathbf{x}-\hat{%
\z}^j)^T \hat{\Sigma}^{-1}(\mathbf{x}-\hat{%
\z}^j),  \label{eq:discri4}
\end{equation}
where the \emph{pooled} covariance matrix estimator is given by 
\begin{equation}
\hat{\Sigma} \,=\, \frac{\sum_{j=0}^{c-1} (n^j-1) \hat{\Sigma}^j}{n-2}\,.
\end{equation}
The LDA classifier is then given by 
\begin{equation}
\psi_{\mathrm{LDA}}\!(\mathbf{x}) \,=\, \operatornamewithlimits{argmax}%
_{j=0,1,\ldots,c-1} D_{\mathrm{LDA}}^j(\mathbf{x})\,.
\end{equation}
The QDA and LDA decision boundaries are composed of pieces of hyperquadric
surfaces and hyperplanes; see \cite{DudaHartStor:01} for more details.

The \emph{naive} QDA and LDA classification rules ignore the nonstationarity
in the data and plug in the usual sample means and sample covariance
matrices based on all the data in the previous formulas.
By contrast, we are in position to define (SMC-based) nonstationary LDA and
QDA (NSLDA and NSQDA, for short) classifiers at each time point $k$, which 
\emph{also} use the entire data, but are adapted to the state of the
evolving distribution at time $k$. This is done by defining discriminants 
\begin{equation}
D_{\mathrm{QDA},k}^{j}\!(\mathbf{x})=\log \pi ^{j}-\frac{1}{2}\log |\hat{%
\Sigma}_{k}^{j}|-\frac{1}{2}(\mathbf{x}-\hat{\z_{k}}%
^{j})^{T}(\hat{\Sigma}_{k}^{j})^{-1}(\mathbf{x}-\hat{\z%
}_{k}^{j}),  \label{eq:discri3n}
\end{equation}%
and 
\begin{equation}
D_{\mathrm{LDA},k}^{j}\!(\mathbf{x})=\log \pi ^{j}-\frac{1}{2}(\mathbf{x}-%
\hat{\z}_{k}^{j})^{T}\hat{\Sigma}_{k}^{-1}(\mathbf{x}-%
\hat{\z}_{k}^{j}),  \label{eq:discri4n}
\end{equation}%
with 
\begin{equation}
\hat{\Sigma}_{k}\,=\,\frac{\sum_{j=0}^{c-1}(n_{k}^{j}-1)\hat{\Sigma}_{k}^{j}%
}{n_{k}-2}\,,
\end{equation}%
and defining the classifiers 
\begin{equation}
\begin{aligned} \psi_{{\rm QDA},k}({\bf x}) & \,=\,
\operatornamewithlimits{argmax}_{j=0,1,\ldots,c-1} D_{{\rm QDA},k}^j\!({\bf
x})\,,\\ \psi_{{\rm LDA},k}({\bf x}) & \,=\,
\operatornamewithlimits{argmax}_{j=0,1,\ldots,c-1} D_{{\rm LDA},k}^j\!({\bf
x})\,. \end{aligned}
\end{equation}%
for $k=1,\ldots ,T$.

The entire process for the proposed SMC-based nonstationary discriminant
analysis, for both the QDA and LDA cases, is summarized in Fig.~\ref%
{fig:diag}. % and Algorithm \ref{alg:NSLDA}.

\begin{figure}[h!]
\centering
\includegraphics[scale=0.31]{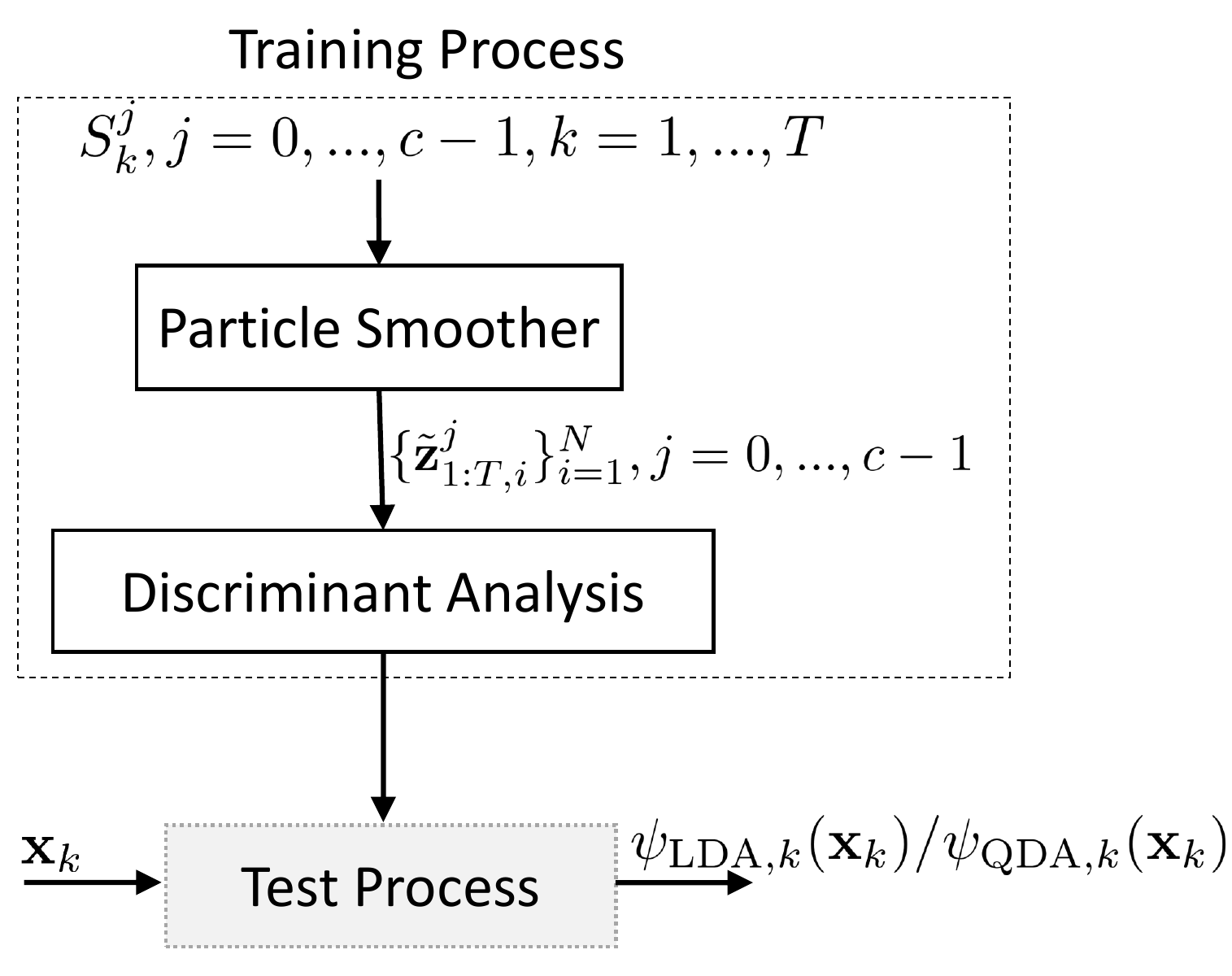}
\caption{Proposed SMC-based Nonstationary Discriminant Analysis
	classification algorithm.}
\label{fig:diag}
\end{figure}

\begin{table}[h!] 	
	\renewcommand{\arraystretch}{1.3}
	\par
	\begin{center}
		
		\begin{tabularx}{\columnwidth}{p{2.6cm}p{5.8cm}}
			\hline
			Parameters &   Value    \\  [1ex]    \hline
			
			Total time points & T = 4, 6, 8, 10, 12 \\ 
			
			Sample size &  n = 20  \\
			
			Dimensionality & d = 2 \\  
			
			Initial means & $\mu_0^0=\begin{bmatrix}1.2  \\ 0.5 \end{bmatrix}, 
			\mu_0^1= \begin{bmatrix} 0.2  \\ 1.5 \end{bmatrix}$ \\  [1.5ex]   
			
			Initial covariances  &  $P_0^0= 0.1I_d, P_0^1= 0.1I_d $\\ [1.5ex]   
			
			\multirow{4}{4em}{Evolution matrix}  & $A^0 = \begin{bmatrix}1.3 & 0.2  \\ 0.9 & 0.6 \end{bmatrix}, 
			A^1 = \begin{bmatrix}0.9 & 0.8  \\ 1 & 0.5 \end{bmatrix} $\\ [1.5ex]    
			
			&  $C^0 = I_d, C^1 = I_d , B^0 = I_d$, \\ 
			& $B^1 = I_d, D^0 = I_d$, $D^1 = I_d$ \\ 
			
			\multirow{2}{4em}{Noise} & $Q^0 = 0.1 I_d, Q^1 = 0.1 I_d $, \\   
			& $R^1 = 0.2 I_d, R^1 = 0.2 I_d $ \\  [1ex]   
			
			\hline
			
		\end{tabularx}
	\end{center}
	\caption{Parameter Settings for case in \ref{sub:linear-example}}
	\label{Table: par1}
\end{table}

\section{Results and Discussion}
We evaluate the proposed nonstationary discriminant analysis (NSDA) under both linear and nonlinear drift. In all experiments, classifiers are trained using all labeled data and evaluated at each target time index; reported errors are Monte Carlo averages over large synthetic test sets (1000 runs unless noted). Baselines are \emph{time–agnostic} rules that pool observations across time~\cite{gama2014conceptdrift}: (i) naive LDA and (ii) a naive RBF–SVM (for nonlinear settings). Our methods produce \emph{time–indexed} discriminants by estimating class moments via smoothing while borrowing information across all time points.

\subsection{Linear Drift Experiments}
\label{sub:linear-example}
We consider the linear–Gaussian model of Eqs.~\eqref{eq:state-linear} with parameters in Table~\ref{Table: par1} (feature dimension $d=2$; total samples $n=20$; horizons $T\in\{3,4,5,6,7,8,9,10\}$). Two scenarios are examined. In \emph{Case One}, the initial states and dynamics matrices $A^j$ are unknown and are estimated via the EM–based Kalman smoother (initialized as in Table~\ref{Table: guess_linearcase}). In \emph{Case Two}, in addition to unknown $z_0^j$ and $A^j$, the measurement time labels are unobserved; we recover both labels and parameters using the GMM–based Kalman smoother (transition noise $Q^j$ known). In both cases, the smoothed trajectories yield time–$k$ class means and covariances for NSLDA; naive LDA estimates static moments from all pooled data.
\begin{table}[h!] 
	\caption{Parameter Settings for case in \ref{sub:linear-example}}
	\renewcommand{\arraystretch}{1.3}
	\par
	\begin{center}
		
		\begin{tabularx}{\columnwidth}{p{2cm}p{5.8cm}}
			\hline 
			Parameters &   Value    \\      \hline  
			
			Initial means & $\mu_{0, \text{guess}}^0=\begin{bmatrix}1  \\ 1 \end{bmatrix}, 
			\mu_{1, \text{guess}}^1= \begin{bmatrix} 1  \\ 1 \end{bmatrix}$ \\  [1.5ex]   
			
			\multirow{2}{4em}{Matrix $A$}  & $A_{\text{guess}}^0 = \begin{bmatrix}1 & 1.5  \\ 0.8 & 1.2 \end{bmatrix}, 
			A_{\text{guess}}^1 = \begin{bmatrix}0.8 & 1.3  \\ 0.5 & 2 \end{bmatrix} $\\ [1.5ex]    
			
			\hline
			
		\end{tabularx}
	\end{center}
	\label{Table: guess_linearcase}
\end{table}

\begin{figure}[t!]
\centering
\includegraphics[scale=0.45]{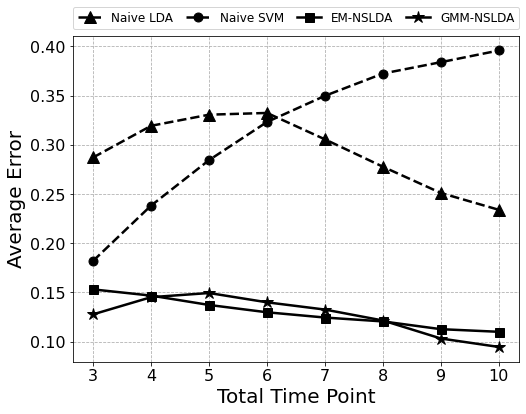}
\caption{Linear drift: average classification error versus horizon $T$ (1000 runs). Dashed curves are pooled baselines—naive LDA ($\blacktriangle$) and naive RBF–SVM ($\bullet$)—that ignore temporal evolution. Solid curves are NSLDA with EM–Kalman smoothing (Case One, $\blacksquare$) and NSLDA with GMM–Kalman smoothing (Case Two, $\bigstar$). NSLDA consistently attains the lowest error, and its advantage increases with $T$ as pooled baselines become increasingly misaligned with the drifting distributions.}
\label{fig:err_linear_example}
\end{figure}

\paragraph*{Performance.}
Figure~\ref{fig:err_linear_example} summarizes average error versus horizon. Both NSLDA variants (EM–NSLDA for Case One; GMM–NSLDA for Case Two) achieve the lowest error across all $T$. The gap over naive LDA widens as $T$ increases, reflecting the growing mismatch of pooled, time-invariant rules under drift. The pooled SVM, while flexible in feature space, remains inferior because it ignores temporal structure. These trends show that accurate state estimation—either through EM parameter learning or through joint time-label recovery—directly translates into better time-indexed classification.

\subsection{Nonlinear Drift Experiments}
\label{sub:nonlinear-example}
We next study a two–class, two–dimensional nonlinear drift (Eq.~\eqref{eq:state}) with additive Gaussian observation noise (Eq.~\eqref{eq:obs2}). Unless stated otherwise, priors are balanced, $T=6$, and initial states are $z_0^0\!\sim\!\mathcal N([1,2]^T,I_2)$ and $z_0^1\!\sim\!\mathcal N([1,3]^T,I_2)$. Latent trajectories are inferred with the particle smoother (Section~\ref{sub:particle-smoother}), after which time–indexed moments are plugged into NSLDA (Section~\ref{sub:SMC-NSLDA}).

\begin{figure*}[h!]
\centering
\includegraphics[scale=0.38]{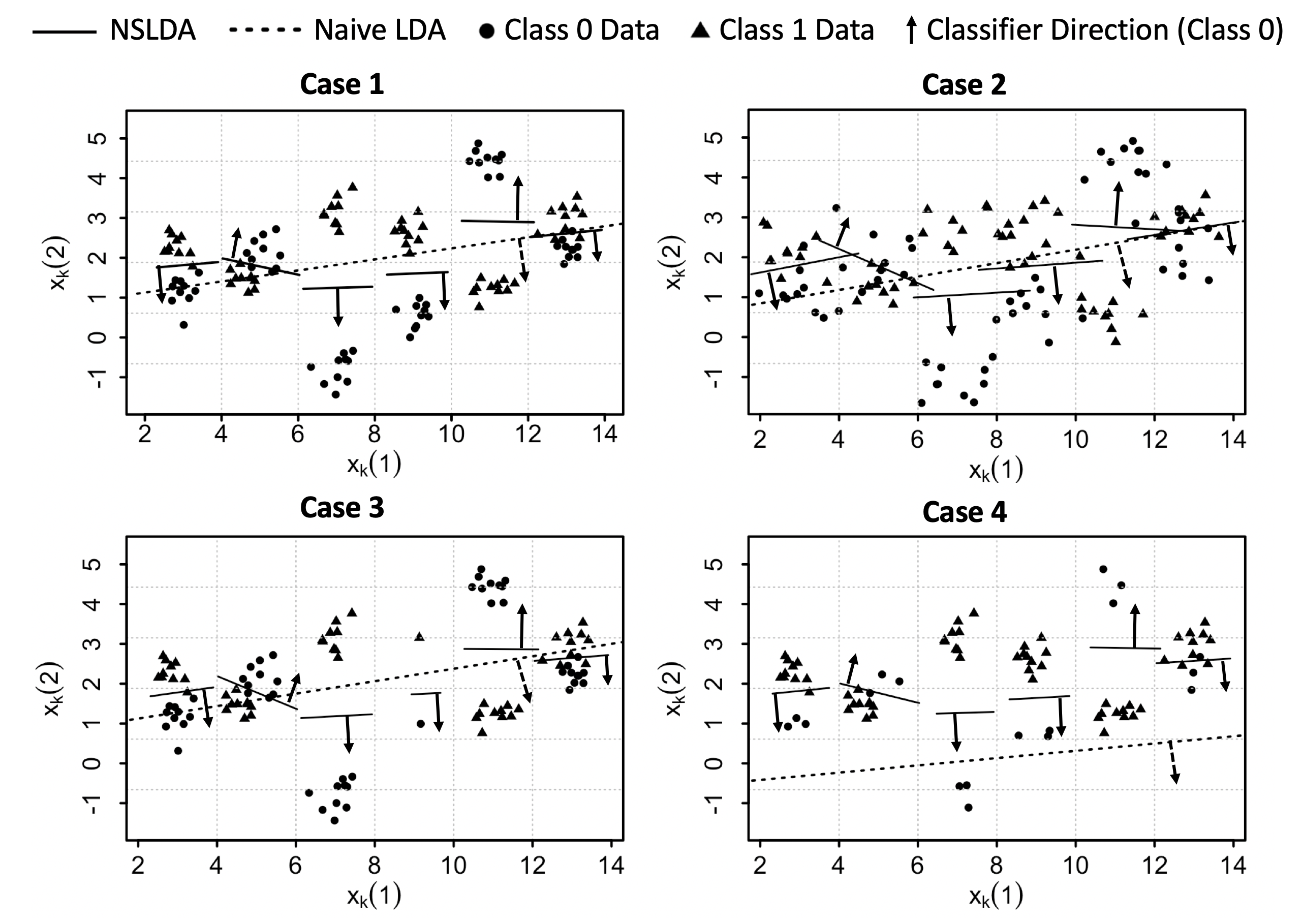}\vspace{-3ex}
\caption{Nonlinear drift (time–indexed boundaries). Naive LDA (dashed) pools across time and produces misaligned or biased boundaries under: (1) nonlinear separation, (2) higher noise, (3) missing data at $k{=}4$, and (4) class imbalance. SMC–based NSLDA (solid) uses smoothed state estimates to tailor a boundary at each time index, yielding stable and well-aligned separation across all stressors (arrows indicate Class~0 direction).}
\label{fig:diag4}
\end{figure*}

\paragraph*{Decision boundaries under data stressors.}
Figure~\ref{fig:diag4} visualizes boundaries for four practically relevant conditions: (1) balanced sampling with low noise ($R^j=0.01I_2$), (2) higher noise ($R^j=0.1I_2$), (3) severe sample scarcity at a single time ($n_4^0=n_4^1=1$), and (4) class imbalance ($n_k^0=10$, $n_k^1=3$). Naive LDA (dashed) underfits the nonlinear separation (Case~1), degrades markedly under higher noise (Case~2), collapses when data are missing at the target time (Case~3), and shifts toward the majority class (Case~4). In contrast, SMC–based NSLDA (solid) adapts the boundary at each time by borrowing strength across the entire trajectory, remaining stable under noise, sparsity, and imbalance.

\begin{figure}[h!]
\centering
\includegraphics[scale=0.42]{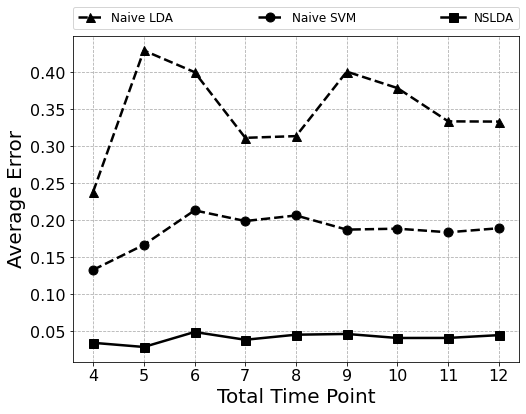}
\caption{Nonlinear drift: average misclassification error versus horizon $T$ (1000 runs, balanced sampling, low noise). Pooled baselines—naive LDA ($\blacktriangle$) and naive RBF–SVM ($\bullet$)—ignore temporal evolution and incur higher error. NSLDA (solid $\blacksquare$) leverages smoothing to produce time–indexed discriminants and maintains the lowest error across all horizons.}
\label{fig:result_err3}
\end{figure}

\paragraph*{Average error versus horizon.}
We then compare naive LDA, naive RBF–SVM, and NSLDA over horizons $T\in\{4,6,8,10,12\}$ with balanced sampling ($n_k^0=n_k^1=10$) and low noise ($R^j=0.01I_2$). Figure~\ref{fig:result_err3} shows that naive LDA has the highest error due to nonlinearity; naive SVM improves by allowing nonlinear boundaries yet still ignores temporal drift; NSLDA achieves the lowest error throughout by smoothing across time while targeting the decision rule to each $k$.

\subsection{Summary of Findings}
Across linear and nonlinear drift, three conclusions emerge. First, \emph{state estimation quality is paramount}: EM learning, GMM-based time-label recovery, and particle smoothing all produce accurate trajectories whose moments yield strong discriminants. Second, \emph{borrowing strength across time} makes the method robust to noise, missing data at specific times, and substantial class imbalance. Third, \emph{time–indexed classification dominates pooled baselines}: even a flexible pooled SVM underperforms a well-estimated, time-specific discriminant that respects temporal dynamics. Together, these results demonstrate that a state–space view of discriminant analysis delivers consistent accuracy gains under temporal distribution shift without resorting to ad-hoc windowing or threshold heuristics.

% \section{Conclusion}
% In this paper, we discuss the state-space approach to nonstationary discriminant analysis. We address the case where both parameters and time points of the measurements are unknown, by using a combination of Gaussian mixture models (GMM) and classical Kalman Smoother (KS) to estimate the time labels, and then using these values with a Kalman smoother (KS) for estimating the unknown parameters. Furthermore, we proposed a general nonlinear, non-Gaussian model for nonstationary data, which allowed us to derive non-stationary discriminant analysis classification rules capable of producing classifiers tuned to the state of the distribution at each time point, while borrowing information from all time points. The proposed framework uses the sequential Monte- Carlo (SMC) estimation of the class conditional density centroids at all time points using all available data. The high accuracy of the proposed NSLDA classification rule and its ability in handling missing or unbalanced data is demonstrated in a series of numerical experiments.

\section{Conclusion}
We presented a state–space approach to nonstationary discriminant analysis. In the linear–Gaussian setting, we addressed incomplete knowledge by combining EM–based Kalman smoothing (with unknown initial states and dynamics) with a GMM–Kalman procedure that also recovers missing
time labels. For general nonlinear or non-Gaussian drift, we developed an SMC smoothing framework (with particle–EM when parameters are unknown) to estimate time–indexed class moments. Plugging these smoothed moments into standard LDA/QDA yields \emph{time-targeted} discriminants that use all observations while adapting the decision rule to any prediction time. Simulations demonstrate consistent error reductions relative to pooled, time-agnostic baselines (naive LDA and a nonlinear SVM) and show robustness to noise, missing or uncertain timestamps, and class imbalance—without ad-hoc windowing.

Future work includes establishing risk guarantees for the time-indexed classifiers (e.g., excess-risk bounds from smoothed-moment error), identifiability and convergence conditions for the EM/particle–EM procedures, and guidance on regularization for small-sample regimes. In addition, on the computational side, the methodology might benefit from faster smoothers (e.g., Rao–Blackwellization, parallel SMC, backward-simulation variants) and automatic model selection for drift dynamics and observation noise. Extensions to richer class models (time-varying covariances or mixtures), semi-supervised label–time recovery, and applications on real, large-scale datasets are also promising directions.

\bibliographystyle{IEEEtran}
\bibliography{refs}

\end{document}